\pdfoutput=1
\documentclass[11pt]{article}
\usepackage{arxiv}
\usepackage{tikz}
\usetikzlibrary{arrows.meta, positioning, shadows.blur, calc}

\usepackage{times}
\usepackage[T1]{fontenc}
\usepackage[utf8]{inputenc}
\usepackage{multirow}
\usepackage{enumitem}
\usepackage{xspace}
\usepackage{booktabs} 

\usepackage{graphicx}
\usepackage{amsmath,amssymb,amsfonts}
\usepackage[title]{appendix}
\usepackage{xcolor}
\usepackage[mathscr]{euscript}
\usepackage{textcomp}
\usepackage{manyfoot}
\usepackage{algorithm}
\usepackage{algorithmicx}
\usepackage{algpseudocode}
\usepackage{listings}
\usepackage[numbers]{natbib}
\usepackage[colorlinks]{hyperref}
\hypersetup{bookmarksdepth=subsection}

\hypersetup{
pdftitle={Building Large-Scale English-Romanian Literary Translation Resources with Open Models},
pdfauthor={Mihai Nadăș, Laura Dioșan, Andreea Tomescu, Andrei Pișcoran},
pdfkeywords={machine translation, literary translation, low-resource languages, Romanian, LLM},
bookmarksnumbered,
pdfstartview={FitH},
colorlinks=true,
linkcolor=blue,
citecolor=blue,
urlcolor=blue,
}

\begin{document}

\title{Building Large-Scale English–Romanian Literary Translation Resources with Open Models}

\author{
  Mihai Nadăș\textsuperscript{1} \and
  Laura Dioșan\textsuperscript{1} \and
  Andreea Tomescu\textsuperscript{1,2} \and
  Andrei Pișcoran\textsuperscript{1,2} \\[1ex]
  \textsuperscript{1}Babeș-Bolyai University, Cluj-Napoca, Romania \\
  \textsuperscript{2}KlusAI Labs, Cluj-Napoca, Romania \\[1ex]
  \texttt{\{mihai.nadas, laura.diosan\}@ubbcluj.ro} \\
  \texttt{\{andreea.tomescu, andrei.piscoran\}@klusai.com}
}


\newcommand{\tinyfabulistframework}{\textsc{TinyFabulist Framework (TF)}\xspace}
\newcommand{\tinyfabulistframeworkshort}{\textsc{TF}\xspace}

\newcommand{\tinyfabulisttranslationframework}{\textsc{TinyFabulist Translation Framework (TF2)}\xspace}
\newcommand{\tinyfabulisttranslationframeworkshort}{\textsc{TF2}\xspace}

\newcommand{\tfoneenThreeM}{\textsc{DS-TF1-EN-3M}\xspace}       
\newcommand{\tftwoenroThreeM}{\textsc{DS-TF2-EN-RO-3M}\xspace}  
\newcommand{\tftwoenroFifteenK}{\textsc{DS-TF2-EN-RO-15K}\xspace}

\newcommand{\tftwoTwelveB}{\textsc{TF2-12B}\xspace}             

\maketitle

\begin{abstract}

Literary translation has recently gained attention as a distinct and complex task in machine translation research, yet translation by small open models remains an open problem, particularly for low-resource languages such as Romanian. We introduce \tinyfabulisttranslationframework, a unified framework for dataset creation, fine-tuning, and evaluation in English$\rightarrow$Romanian literary translation. Building on \tfoneenThreeM, the largest collection of synthetic English fables to date, our pipeline first generates 15k high-quality Romanian references from the TF1 pool using a high-performing large language model (LLM). We then apply a two-stage fine-tuning process to a 12B-parameter open-weight model: (i) instruction tuning to capture genre-specific narrative style, and (ii) adapter compression for efficient deployment. Evaluation combines a five-dimension LLM-based rubric (accuracy, fluency, coherence, style, cultural adaptation) as the primary comparative framework, alongside corpus-level Bilingual Evaluation Understudy (BLEU) reported as a secondary reference-based consistency metric. Our fine-tuned model (\tftwoTwelveB) achieves strong fluency and adequacy, narrowing the gap to top-performing proprietary models under automated and human-anchored evaluation, while being open, accessible, and significantly more cost-effective. We publicly release the fine-tuned model and two large-scale synthetic parallel datasets (\tftwoenroThreeM and \tftwoenroFifteenK), along with all scripts and evaluation prompts. \tinyfabulisttranslationframeworkshort provides an end-to-end, reproducible pipeline for research on cost-efficient translation, cross-lingual narrative generation, and the broad adoption of open models for culturally significant literary content in low-resource settings.
\end{abstract}

\section{Introduction} 
Translating literature into low-resource languages poses unique challenges, given both the scarcity of parallel data and the need to faithfully preserve stylistic nuance and cultural context. Romanian---a Latin language spoken by over 24 million people---remains notably underserved in the realm of literary translation resources. Existing machine translation (MT) benchmarks, such as WMT\footnote{WMT stands for the Conference on Machine Translation (formerly ``Workshop on Machine Translation''). The WMT organizes annual shared tasks for benchmarking machine translation systems, with its ``news translation'' task being the most widely used standard evaluation for translation of contemporary news articles. See: \url{https://www.statmt.org/wmt23/}.}, predominantly focus on informational or news text, thus providing limited guidance for creative, narrative-driven translation tasks. Recent advances in large language models (LLMs) have enabled synthetic text generation at unprecedented scale, opening new opportunities for augmenting MT datasets \citep{long_llms-driven_2024, nadas_synthetic_2025}. Notably, projects such as TinyStories \citep{eldan_tinystories_2023} demonstrate that even compact models can generate coherent narratives when trained on carefully curated synthetic corpora. However, much of the existing work has focused on \emph{monolingual} story generation in English \citep{guan_corpus_2022, eldan_tinystories_2023} or on improvements to translation via back-translation \citep{sennrich_improving_2016}, rather than on the systematic creation of large-scale bilingual literary datasets. As a result, open questions remain regarding how both proprietary and open-source LLMs can be leveraged to construct and evaluate literary translations in genuinely low-resource settings.

Substantial progress has also been made in building large bilingual and multilingual resources. For example, CLIRMatrix~\citep{sun_clirmatrix_2020} compiles parallel data across 139 languages for cross-lingual retrieval, and mT5~\citep{xue_mt5_2021} pretrains a text-to-text transformer on a wide range of language pairs and domains. These resources, however, primarily target general-domain or retrieval-oriented MT and include limited literary content. Even for language pairs such as English--Romanian, coverage is typically restricted to news, web, or miscellaneous domains. To date, no openly available, large-scale English--Romanian parallel corpus has been specifically curated for creative narratives or fables.

\textbf{\tinyfabulisttranslationframework} addresses this gap by extending our earlier TinyFabulist-TF1 corpus of English-only fables~\citep{nadas_tf1-en-3m_2025} into a bilingual setting. The resulting dataset, \tftwoenroThreeM, comprises three million synthetic English--Romanian parallel fables, accompanied by fine-tuned open-source translation models and a literary translation benchmark. By combining parameter-efficient model adaptation with a narrative-oriented evaluation protocol, \tinyfabulisttranslationframeworkshort provides both a large-scale resource and a practical experimental framework for studying literary machine translation in a low-resource language pair.

Recent multilingual models such as NLLB-200~\citep{team_no_2022} and EuroLLM~\citep{martins_eurollm-9b_2025} achieve strong performance across many language pairs, but are generally optimized for breadth and general-domain translation. Their large parameter counts can also limit deployment in cost-constrained or on-device scenarios. In our experiments, these models do not consistently capture the stylistic nuance, discourse coherence, and cultural adaptation required for literary translation (Section~\ref{sec:results}), motivating the exploration of more targeted, domain-adaptive approaches.

Building on the English-only fable generation efforts of \tfoneenThreeM \citep{nadas_tf1-en-3m_2025}, we focus on moral fables---short narratives that require semantic fidelity alongside fluency, narrative coherence, stylistic faithfulness, and culturally appropriate moral framing. Generating parallel data for such content under resource constraints motivates the use of LLMs for both dataset construction and scalable evaluation.

Our contributions are summarized as follows:
\begin{itemize}
    \item \textbf{Narrative-Aware Evaluation Protocol.}  
    We introduce a multi-dimensional evaluation framework for literary translation that combines corpus-level BLEU \citep{sulem_bleu_2018} with an LLM-based rubric assessing accuracy, fluency, coherence, style, and cultural adaptation. In this work, BLEU is used strictly as a secondary consistency check, while primary analysis relies on rubric-based LLM evaluation with cross-family bias validation.

    \item \textbf{Two openly licensed corpora for literary MT.}  
    We release (i) \tftwoenroFifteenK, a 15\,000-example English$\rightarrow$Romanian parallel corpus of moral fables, and (ii) \tftwoenroThreeM, a three-million-example corpus automatically translated using fine-tuned open models.\footnote{\url{https://huggingface.co/datasets/klusai/tf2-en-ro-15k},\;%
    \url{https://huggingface.co/datasets/klusai/tf2-en-ro-3m}}

    \item \textbf{Three open, fine-tuned translation models.}  
    We release three LoRA-adapted English--Romanian translation models---\texttt{tf2-1b}, \texttt{tf2-4b}, and \texttt{tf2-12b}---which substantially narrow the performance gap to proprietary systems under cost-aware evaluation, while remaining efficient enough to run on commodity GPUs.\footnote{\url{https://huggingface.co/klusai/tf2-1b},\;
    \url{https://huggingface.co/klusai/tf2-4b},\;
    \url{https://huggingface.co/klusai/tf2-12b}}

    \item \textbf{Fully transparent artifacts.}  
    All datasets, code, prompts, and evaluation scripts are released under permissive licenses to support reproducibility and further research in low-resource literary NLP.
\end{itemize}

\subsection{Research Questions}

The construction of large-scale translation datasets and models is constrained by multiple practical considerations. Financial cost is often the most immediate limitation, particularly for low-resource language communities or research teams with restricted budgets. Additional factors include the environmental impact of large-scale computation---such as CO$_2$ emissions and water use for hardware cooling---and the ethical provenance of training data, especially with respect to data privacy and fair representation. These considerations motivate the design of our experimental pipeline and inform the trade-offs examined in this work.

Accordingly, we structure our study around the following research questions:
\begin{enumerate}
    \item[\textbf{RQ1}] \textit{Cost-Constrained Data Generation}: Can a large-scale, high-quality English--Romanian literary translation dataset be built using small open models under strict budget constraints, and what trade-offs does this entail?
    \item[\textbf{RQ2}] \textit{Open vs. Proprietary Translation Quality}: How do fine-tuned open models compare to proprietary systems for English--Romanian literary translation under cost-aware and resource-constrained settings?
    \item[\textbf{RQ3}] \textit{Evaluation Rigor and Automation}: To what extent can an automated, LLM-based evaluation framework provide useful and interpretable signals for literary translation quality?
\end{enumerate}

The remainder of this paper is organized as follows. Section~\ref{sec:related} reviews related work on literary machine translation, synthetic data generation, and evaluation. Section~\ref{sec:framework} describes the TF2 pipeline in detail. Section~\ref{sec:results} presents experimental results and comparative evaluations. Section~\ref{sec:tf2-dataset} introduces the released datasets and metadata schema. Sections~\ref{sec:discussion} and~\ref{sec:threats} discuss findings and limitations, respectively.

\section{Related Work}
\label{sec:related}

\paragraph{Synthetic data and low-resource Machine Translation (MT).}
Large language models have been widely used to generate training corpora for tasks where human-annotated data are scarce or costly. Early work on back-translation \citep{sennrich_improving_2016} demonstrated that automatically generated target--source pairs can substantially improve neural MT in low-resource settings. More recent LLM-driven pipelines have enabled the creation of massive monolingual and parallel datasets at a fraction of traditional annotation cost \citep{long_llms-driven_2024, wang_self-instruct_2023_2}. In narrative domains, \citet{eldan_tinystories_2023} introduced \textsc{TinyStories}, showing that large collections of synthetic narratives can support coherent story generation even with compact models. Despite these advances, synthetic data for bilingual literary translation remain comparatively underexplored.

\paragraph{Fable generation and moral reasoning.}
Moral and didactic narratives have attracted interest as test-beds for studying value-aligned generation and commonsense reasoning. \citet{guan_corpus_2022} introduced \textsc{STORAL}, a human-authored dataset pairing narratives with explicit moral lessons, which has since been used for tasks such as moral inference and story understanding. Subsequent work has explored scaling narrative datasets and using synthetic generation to model moral reasoning in text. These efforts highlight the suitability of short narrative forms, such as fables, for studying higher-level discourse phenomena, including coherence, moral framing, and cultural interpretation.

\paragraph{Literary machine translation.}
Literary translation has received increasing attention as a distinct MT challenge, requiring not only semantic accuracy but also preservation of style, tone, and cultural nuance. The WMT 2023 shared task on discourse-level literary translation~\citep{wang_findings_2023} highlighted the difficulty of maintaining coherence across longer narrative spans. \citet{karpinska_large_2023} demonstrated that while LLMs can effectively leverage document-level context for literary translation, critical errors---such as mistranslations, omissions, and inconsistent character references---persist even in state-of-the-art systems. More recently, \citet{zhang_how_2025} systematically compared LLM-based and human evaluation for literary translation, finding that automatic metrics often favor literal renderings over stylistically appropriate translations, underscoring the need for careful interpretation of automated scores in this domain.

\paragraph{Open vs.\ proprietary LLMs for translation.}
Comparative studies of proprietary APIs (e.g., GPT-4, Gemini) and open-weight models (e.g., LLaMA, Mistral, Qwen) report a persistent but narrowing quality gap, particularly for less-represented languages and domains~\citep{zheng_judging_2023}. Parameter-efficient fine-tuning methods, most notably LoRA adapters~\citep{hu_lora_2021} and subsequent PEFT toolkits~\citep{weyssow_exploring_2025}, have been shown to substantially improve task-specific performance of open models while keeping training and deployment costs low. Recent multilingual open-weight models such as \textbf{DeepSeek-LLM}~\citep{deepseek-ai_deepseek_2024}, \textbf{EuroLLM-9B}~\citep{martins_eurollm-9b_2025}, and Meta’s \textbf{NLLB-200}~\citep{team_no_2022} demonstrate strong general-purpose translation performance across many languages, including Romanian. However, their scale and hardware requirements can limit accessibility in cost- or resource-constrained settings, motivating continued interest in compact, adaptable translation models.

\paragraph{Evaluation beyond BLEU.}
BLEU \citep{papineni_bleu_2002} remains a widely used reference-based metric for MT evaluation, but its correlation with human judgment is known to decrease for creative and stylistically rich text. Alternative frameworks such as MQM \citep{lommel_multidimensional_2014} and COMET \citep{rei_comet_2020} provide more fine-grained assessments of adequacy and fluency, though they typically require expert annotations or supervised quality estimation models. More recently, LLMs have been employed as automatic judges, showing promising agreement with human evaluations in tasks such as summarization and dialogue assessment \citep{liu_g-eval_2023}. Notably, \citet{kocmi_large_2023} demonstrate that LLM-based evaluators can achieve state-of-the-art correlation with human judgments specifically for machine translation, motivating the use of LLM judges in MT pipelines. However, LLM judges are known to exhibit systematic biases, including self-preference bias---the tendency to favor outputs from similar model families~\citep{wataoka_selfpref_2024}---and position bias in pairwise comparisons~\citep{zheng_judging_2023}. These findings underscore the importance of multi-model evaluation and cross-family validation when deploying LLM-based assessment, practices we adopt in our evaluation pipeline (Section~\ref{sec:results}).

Several recent approaches further extend evaluation in directions relevant to narrative translation. \citet{hu_exploring_2023} propose \textsc{Cont-COMET}, a context-aware extension of COMET that incorporates surrounding sentences to better capture discourse-level effects. \citet{wang_maats_2025} introduce a multi-agent evaluation framework in which multiple LLMs assess terminology consistency, narrative perspective, and stylistic fidelity. \citet{zhang_litransproqa_2025} present \textsc{LiTransProQA}, a question-answering-based evaluation method targeting style, tone, and cultural references. Finally, \citet{yao_benchmarking_2024} release the CAMT corpus alongside metrics designed to evaluate culture-specific translation phenomena, demonstrating the importance of culturally aware evaluation signals.

\paragraph{Cost-aware NLP pipelines.}
A growing body of work emphasizes budget-constrained model development and deployment, ranging from efficient inference and quantization techniques \citep{frantar_gptq_2023} to dataset generation strategies that trade limited API usage for downstream performance gains \citep{wang_reasoning_2024}. These efforts illustrate how careful design choices can enable competitive NLP systems under practical cost and hardware constraints.

\paragraph{Open-source Romanian language resources.}
Recent initiatives have begun to address the scarcity of open benchmarks and models for Romanian NLP. \citet{masala_vorbesti_2024} introduce \textsc{OpenLLM-Ro}, a collection of open-source Romanian language models, datasets, and evaluation benchmarks released via GitHub\footnote{\url{https://github.com/openllm-ro}} and Hugging Face\footnote{\url{https://huggingface.co/OpenLLM-Ro}}. While these resources provide a valuable foundation for Romanian language technology, they primarily target general-purpose language understanding and generation rather than literary translation or creative narrative domains.

\vspace{0.5em}
Taken together, prior work highlights the potential of synthetic data, parameter-efficient adaptation, and cost-aware evaluation strategies for advancing machine translation in low-resource settings. These insights motivate the pipeline and resources introduced in the following sections.

\section{\tinyfabulisttranslationframework}
\label{sec:framework}

\tinyfabulisttranslationframeworkshort generalizes the original \tinyfabulistframework{} project to a bilingual and scalable setting, introducing an end-to-end methodology for constructing an English--Romanian literary translation resource. The framework emphasizes transparency, replicability, and cost-awareness, prioritizing open artifacts, parameter-efficient adaptation, and evaluation that combines automatic metrics, rubric-based assessment, and targeted human judgment.

\begin{figure}[!htbp]
\centering
\resizebox{\textwidth}{!}{%
\begin{tikzpicture}[
  >=Stealth,
  node distance=0.5cm and 1.1cm,
  stage/.style={
    draw, very thick, rounded corners=3pt,
    minimum height=1.9cm, minimum width=3.2cm,
    align=left, font=\footnotesize,
    top color=white, bottom color=blue!10,
    drop shadow={blur shadow}
  }]
  \node[stage] (select) {
    \textbf{1. Evaluate \& Benchmark}\\
    > 13 LLM/MT candidates\\
    > literary metrics
  };
  \node[stage, right=of select] (create) {
    \shortstack[l]{
      \textbf{2. Create 15k Parallel Corpus}\\
      > EN$\rightarrow$RO (best system)\\
      > reference-quality split
    }
  };
  \node[stage, right=of create] (finetune) {
    \shortstack[l]{
      \textbf{3. Fine-Tune Open LLMs}\\
      > 1B / 4B / 12B backbones\\
      > LoRA, quantized, reproducible
    }
  };
  \node[stage, right=of finetune] (mass) {
    \shortstack[l]{
      \textbf{4. Mass Translation}\\
      > $\sim$3M English fables\\
      > Romanian translations
    }
  };
  \draw[very thick,->] (select.east) -- (create.west);
  \draw[very thick,->] (create.east) -- (finetune.west);
  \draw[very thick,->] (finetune.east) -- (mass.west);
  \path (finetune.north) ++(0,0.32) coordinate (topA);
  \path (select.north) ++(0,0.32) coordinate (topB);
\end{tikzpicture}
}
\caption{\tinyfabulisttranslationframework pipeline: evaluation of translation models on literary benchmarks (S1); creation of a 15k English--Romanian parallel corpus via the top-ranked system (S2); parameter-efficient fine-tuning of open LLMs and quantized variants (S3); and large-scale translation of the full English corpus (S4).}
\label{fig:tf2-pipeline}
\end{figure}
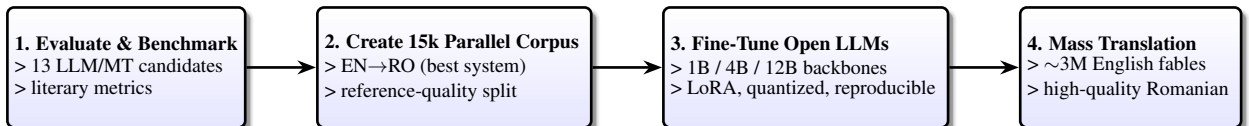

The workflow (see Figure~\ref{fig:tf2-pipeline}) consists of four stages:

\begin{enumerate}[nosep,label=\textbf{S\arabic*}]
    \item \textbf{Evaluating Candidate Translators:}
    We benchmark 13 diverse LLMs and commercial translation APIs on a held-out fable set drawn from \tfoneenThreeM \citep{nadas_tf1-en-3m_2025}. Evaluation targets multiple aspects of literary translation and uses a five-dimension LLM-based rubric to compare systems in a scalable and consistent manner.
    
    \item \textbf{Parallel Dataset Creation:}
    The top-ranked Stage~1 system is used to translate 15{,}000 English fables, producing the TinyFabulist \tftwoenroFifteenK corpus. This silver-standard parallel dataset provides supervision for downstream adaptation in a setting without human-authored references, and supplies a consistent reference corpus for subsequent reference-based evaluation (e.g., BLEU).
    
    \item \textbf{Parameter-Efficient Fine-Tuning:}
    A suite of open LLMs---ranging from 1B to 12B parameters---are fine-tuned on the 15k parallel set using Low-Rank Adaptation (LoRA; \citealp{hu_lora_2021}), yielding domain-adapted English--Romanian translation models. Quantized variants are produced to support efficient local inference.
    
    \item \textbf{Large-Scale Corpus Generation:}
    The best-performing fine-tuned model(s) are used to translate the remaining $\sim$3M English fables from \tfoneenThreeM, resulting in a large bilingual English--Romanian corpus for Romanian literary MT. The pipeline is modular and supports future re-generation as translation models or evaluation protocols improve.
\end{enumerate}

\vspace{.3em}
Each stage is integrated with quality control and open artifact release, providing a reproducible blueprint for low-resource literary MT beyond English--Romanian.

\subsection{Stage 1: Evaluating Models \& Selecting Translators}
\label{sec:eval}

The first stage serves two objectives: (i) selecting a translation system to act as a reference translator for parallel corpus construction, and (ii) identifying a competitive open-source backbone suitable for downstream parameter-efficient fine-tuning. We evaluate a diverse pool of candidate systems, including commercial MT APIs, proprietary LLMs, and open instruction-tuned models, with a focus on English--Romanian literary translation.

\noindent\textbf{Evaluation Methodology.}
Translation quality in literary narratives involves multiple complementary aspects, including semantic fidelity, linguistic well-formedness, discourse coherence, stylistic appropriateness, and cultural adaptation. To capture these facets in a scalable and comparable manner, we employ rubric-based LLM evaluation at Stage~1 and introduce reference-based metrics (e.g., BLEU) once a consistent reference corpus becomes available in later stages.

At Stage~1, where no human-authored Romanian references are available, candidate systems are compared primarily using a multi-dimensional rubric implemented via an LLM acting as an automated judge. This rubric evaluates five dimensions derived from professional translation guidelines:
\begin{enumerate}[nosep]
    \item \textbf{Accuracy:} Fidelity to the semantic content of the source text.
    \item \textbf{Fluency:} Grammaticality, naturalness, and readability in the target language.
    \item \textbf{Coherence:} Logical flow and clarity at the story level.
    \item \textbf{Style:} Appropriateness of tone and consistency with the narrative voice.
    \item \textbf{Cultural/Pragmatic Adaptation:} Appropriateness of cultural references and moral framing for the target audience.
\end{enumerate}

Each dimension is scored on a 1--5 Likert scale (1 = poor, 5 = excellent). To operationalize this rubric at scale, we prompt an LLM to emulate a professional literary translation reviewer:

\emph{"You are a professional translation evaluator. You will be given an English fable and a Romanian translation. Evaluate the translation for accuracy, fluency, coherence, style, and cultural/pragmatic fidelity. Provide a score from 1 to 5 for each category, along with a brief justification. Output your evaluation in valid JSON format with fields for each score and justification."}

\noindent Each dimension is defined explicitly in the prompt using the descriptions listed above, ensuring consistent interpretation across evaluation runs. The structured JSON output enables reliable aggregation of scores across systems. For Stage~1 evaluation, we use \textbf{GPT-o3-mini} as the automated judge, selected for its strong reasoning capabilities, competitive cost, and consistent performance on multilingual evaluation tasks. For each candidate model, we evaluate a randomly sampled subset of 100 distinct fables, yielding per-dimension scores and short qualitative rationales. The resulting averages provide a comparative view of translation behavior across systems. This approach follows prior work showing that LLM-based rubrics can provide useful comparative signals in text generation evaluation~\citep{liu_g-eval_2023}. In our pipeline, these scores are used as an auxiliary decision signal for model selection rather than as an absolute measure of translation quality.

\noindent\textbf{Selecting Translators and Backbones.}
Based on aggregated rubric scores, the highest-ranked system is selected as the reference translator for parallel corpus construction and downstream benchmarking. Independently, the strongest-performing open-source model is selected as the backbone for parameter-efficient fine-tuning in Stage~3.

\subsection{Stage 2: Parallel Corpus Generation (15k Fables)}
\label{sec:data15k}

A central step in constructing a benchmark for low-resource literary MT is the creation of a parallel dataset tailored to the narrative and stylistic demands of the target genre. In \tinyfabulisttranslationframeworkshort, we address the shortage of English--Romanian literary data by leveraging LLMs to generate a diverse set of story translations, following established practices in synthetic corpus construction~\citep{long_llms-driven_2024, eldan_tinystories_2023}.

We sample 15{,}000 short fables from \tfoneenThreeM\footnote{\url{https://huggingface.co/datasets/klusai/ds-tf1-en-3m}}, each consisting of a concise narrative ending with an explicit moral. We then translate them into Romanian using the top-ranked translator selected in Stage~1. This yields a machine-generated parallel corpus intended to serve as both a training resource and a consistent reference set for subsequent benchmarking.

The resulting TinyFabulist \tftwoenroFifteenK dataset is released on Hugging Face\footnote{\url{https://huggingface.co/datasets/klusai/tf2-en-ro-15k}} under the permissive \textbf{MIT License}. Each record is stored in JSONL format and includes the English source, the Romanian translation, and metadata to support downstream analysis and reproducibility. Key fields include:
\begin{itemize}[nosep]
    \item \texttt{fable}: The original English fable text (typically 1--3 paragraphs), ending with an explicit moral.
    \item \texttt{translated\_fable}: The Romanian translation generated by the selected model.
    \item \texttt{pipeline\_stage}: Stage of the data-generation pipeline that produced the record (here \texttt{translation}).
    \item \texttt{source\_lang} and \texttt{target\_lang}: Language codes indicating source and target (here always \texttt{English} $\rightarrow$ \texttt{Romanian}).
    \item \texttt{prompt\_hash}: SHA-256 hash of the translation prompt, used for reproducibility and traceability.
    \item \texttt{llm\_name}: Identifier of the model used for translation (e.g., model name or Hugging Face snapshot).
    \item \texttt{translation\_model}: Identifier of the specific checkpoint used to produce the translation (when distinct from \texttt{llm\_name}).
    \item \texttt{generation\_timestamp}: Unix timestamp indicating when the translation was generated.
\end{itemize}

The dataset is split into 12{,}000 training pairs, 1{,}500 validation pairs, and 1{,}500 test pairs. Although the Romanian references are model-generated, their selection follows the benchmarking procedure described in Section~\ref{sec:eval}. The accompanying metadata supports reproducibility, comparative evaluation, and auditing of translation provenance.

Overall, TinyFabulist \tftwoenroFifteenK provides a transparent foundation for training and evaluating English--Romanian literary MT models in the absence of large human-translated corpora.

\subsection{Stage 3: Fine-Tuning---Parameter-Efficient Domain Adaptation}
\label{sec:finetune}

The third stage adapts open instruction-tuned LLMs to the literary translation domain using parameter-efficient fine-tuning. Instead of updating all model parameters, we employ Low-Rank Adaptation (LoRA; \citealp{hu_lora_2021}), which injects lightweight, trainable adapters into attention and feed-forward layers while keeping most parameters frozen. This approach enables domain adaptation with relatively modest computational overhead.

\noindent\textbf{Training Methodology.}
The fine-tuning process follows established instruction-tuning practices:
\begin{enumerate}[label=(\alph*),leftmargin=1.6em]
    \item \textbf{Data preparation:} Align and preprocess English--Romanian text pairs into instruction--response format, filtering for maximum length (all source fables originate from \tfoneenThreeM, ensuring domain consistency). Each input is prefixed with a standardized prompt (e.g., ``Translate the following fable from English to Romanian:'').
    \item \textbf{Tokenization and masking:} Employ the native SentencePiece or BPE vocabulary associated with each model family. In the \texttt{labels} tensor, mask all tokens except the Romanian output, ensuring that only target-side predictions contribute to the training loss.
    \item \textbf{Adapter injection:} Attach LoRA adapters to major projection matrices in both self-attention and feed-forward blocks---\texttt{q\_proj}, \texttt{k\_proj}, \texttt{v\_proj}, \texttt{o\_proj}, \texttt{gate\_proj}, \texttt{up\_proj}, and \texttt{down\_proj}. For each weight $W\in \mathbb{R}^{d_{\text{out}}\times d_{\text{in}}}$ we learn a low-rank update $\Delta W=\tfrac{\alpha}{r}BA$ with $r{=}32$, $\alpha{=}32$ (effective scaling $=1.0$), and apply dropout $p{=}0.05$ on the adapter path. We train only $A,B$ while freezing the backbone (no bias adaptation), and merge $\Delta W$ into $W$ at inference. This follows best practices in recent literature \citep{mao_survey_2024, ding_lora-c_2024}.
    \item \textbf{Training regime:} Fine-tune models using AdamW optimization, gradient accumulation, cosine learning rate scheduling, and mixed-precision (FP16 or bfloat16) arithmetic. To reduce overfitting, we apply early stopping based on held-out validation loss.
    \item \textbf{Deployment:} After training convergence, merge LoRA adapters into the base weights for standalone inference and produce 8-bit quantized variants \citep{dettmers_gpt3int8_2022, frantar_gptq_2023} to reduce memory and latency. We additionally leverage \texttt{llmcompressor} to generate W8A8 quantized models, compatible with \texttt{vLLM} \citep{kwon_efficient_2023, shaw_llm_2024}. For offline distribution we also export GGUF artifacts compatible with llama.cpp and Hugging Face GGUF endpoints\footnote{\url{https://github.com/ggml-org/llama.cpp}}.
\end{enumerate}

\noindent\textbf{Evaluation.}
Fine-tuned models are assessed using both rubric-based evaluation (as in Stage~1) and BLEU. Here, BLEU is used as a complementary metric by comparing translations against the reference corpus created in Stage~2. Rubric scores remain the primary measure for multi-dimensional translation quality, while BLEU provides a lightweight consistency check on lexical overlap.

\noindent\textbf{Backbone Coverage.}
This fine-tuning procedure is applied across multiple model scales to assess robustness under varying computational constraints. Results are reported in Section~\ref{sec:results}, including changes in both rubric scores and BLEU relative to untuned backbones.

\subsection{Stage 4: Large-Scale Fable Translation}

The final stage translates the full TinyFabulist English corpus into Romanian using the fine-tuned models. \tfoneenThreeM comprises approximately 3 million AI-generated English fables, which we translate into Romanian primarily using the best-performing fine-tuned checkpoint. In addition, for a small number of runs we also generate translations with alternative fine-tuned checkpoints to support ablations and comparative analyses reported in Section~\ref{sec:results}. This process yields the \tftwoenroThreeM dataset, an openly released English--Romanian parallel corpus for literary translation research.

The translation process is designed to be feasible on modest hardware (e.g., small GPU clusters or standard cloud infrastructure) without prohibitive computational cost. The resulting dataset is released on the Hugging Face Hub.\footnote{\url{https://huggingface.co/datasets/klusai/tf2-en-ro-3m}}

\textit{A full statistical and structural overview of the \tftwoenroThreeM dataset---including metadata schema, field definitions, and example use cases---is provided in Section~\ref{sec:tf2-dataset}.}

Complete hardware and software specifications are listed in Appendix~\ref{sec:appendix:costs}.

\section{Experiments and Results}
\label{sec:results}

This section evaluates \tinyfabulisttranslationframeworkshort's translation models against a range of strong baselines---both proprietary and open---using reference-based metrics (BLEU) and an LLM-based rubric. We compare all fine-tuned TF2 models, their quantized variants, and the influence of decoding temperature on translation quality.

\subsection{Benchmarked Systems and Baselines}

We first benchmark only \emph{external} systems as obtained ``out of the box'' from their public endpoints or hubs, with no additional fine-tuning. The pool spans:
\begin{itemize}[leftmargin=1.5em]
    \item \textbf{Proprietary LLMs and commercial MT:}
    GPT-4.1, GPT-4.1-mini, GPT-o3, GPT-o3-mini, Gemini-2.5, Gemini-2.0, Grok-3, DeepL.
    \item \textbf{Open instruction-tuned baselines:}
    EuroLLM-9B, Qwen3-14B, and untuned Gemma-3 models (1B/4B/12B).
\end{itemize}

For each model we compute the per-dimension rubric and the overall average:
\[
\text{Avg.\ Score}=\frac{\text{Accuracy}+\text{Fluency}+\text{Coherence}+\text{Style}+\text{Cultural/Pragmatic}}{5}.
\]

\begin{table*}[!htbp]
\centering\footnotesize
\resizebox{\textwidth}{!}{%
\begin{tabular}{lccccccc}
\toprule
\textbf{Model} & \textbf{Accuracy} & \textbf{Fluency} & \textbf{Coherence} & \textbf{Style} & \textbf{Cultural} & \textbf{Avg.\ Score} & \textbf{Count} \\
\midrule
\textbf{GPT-o3 (2025-04-16)}      & \textbf{4.86} & \textbf{4.92} & \textbf{4.89} & \textbf{4.96} & \textbf{4.97} & \textbf{4.92} & 100 \\
GPT-4.1-mini (2025-04-14)         & 4.54 & 4.71 & 4.72 & 4.84 & 4.83 & 4.73 & 98 \\
GPT-4.1 (2025-04-14)              & \textbf{4.86} & 4.89 & 4.85 & 4.92 & 4.94 & 4.89 & 100 \\
GPT-o3-mini (2025-01-31)          & 4.71 & 4.78 & 4.87 & 4.85 & 4.92 & 4.83 & 100 \\
Gemini-2.5-Flash                  & 4.75 & 4.86 & 4.82 & 4.87 & 4.89 & 4.84 & 100 \\
Gemini-2.0-Flash-001              & 4.66 & 4.82 & 4.78 & 4.89 & 4.93 & 4.82 & 100 \\
Gemini-Flash-1.5-8b               & 4.14 & 4.45 & 4.67 & 4.52 & 4.46 & 4.45 & 99  \\
DeepL                             & 4.42 & 4.73 & 4.38 & 4.69 & 4.74 & 4.59 & 100 \\
Grok-3-mini-beta                  & 4.73 & 4.74 & 4.77 & 4.82 & 4.88 & 4.79 & 100 \\
EuroLLM-9B-Instruct               & 3.84 & 4.27 & 4.36 & 4.27 & 4.22 & 4.19 & 98  \\
Gemma-3-12B-it                    & 3.98 & 4.56 & 4.65 & 4.52 & 4.43 & 4.43 & 100 \\
Gemma-3-4B-it                     & 3.27 & 3.94 & 4.17 & 3.91 & 3.78 & 3.81 & 100 \\
Gemma-3-1B-it                     & 1.79 & 2.13 & 2.23 & 2.07 & 1.86 & 2.02 & 100 \\
\bottomrule
\end{tabular}
}
\caption{Baseline systems (i.e., models evaluated without TF2 fine-tuning) on the literary translation benchmark. Columns are rubric scores (max 5), their mean (Avg.\ Score), and sample count; counts below 100 reflect API failures or malformed JSON responses that could not be parsed. Bold indicates the best value per rubric column. Among open-weight models, Gemma-3-12B-it achieves the highest average (4.43) and is selected as the backbone for TF2 fine-tuning. Each input sequence contains on average 300--400 tokens, with outputs averaging 350--450 tokens.}
\label{tab:baseline-non-tf2}
\end{table*}

\paragraph{Why GPT-o3 as the reference translator?}
Table~\ref{tab:baseline-non-tf2} indicates that GPT-o3 attains the highest average rubric score across the evaluated systems. We therefore adopt GPT-o3 outputs as \emph{silver-standard} references for reference-based comparisons elsewhere in the paper. Subsequent BLEU results use these GPT-o3 translations as references, which is why GPT-o3 itself does not appear with a comparable BLEU entry in those tables.

\paragraph{Why Gemma-3-12B-it as the open backbone to fine-tune?}
Among the open baselines, Gemma-3-12B-it achieves the highest rubric average (4.43), outperforming the 4B (3.81) and 1B (2.02) variants. It also offers a permissive license and mature tooling, making it a practical foundation for adaptation. We therefore select Gemma-3-12B-it (and, for scaling experiments, its 4B and 1B counterparts) as backbones for TF2 fine-tuning.

\subsection{Overall Translation Quality with TF2 Models}
\label{sec:results:tf2}

\paragraph{TF2 fine-tuned models.}
The \tinyfabulisttranslationframeworkshort series comprises three parameter-efficient fine-tuned variants of the \texttt{Gemma-3} instruction-tuned backbones: \texttt{Gemma-3-1B-it}, \texttt{Gemma-3-4B-it}, and \texttt{Gemma-3-12B-it}. Each model is adapted on the \tftwoenroFifteenK corpus using LoRA adapters injected into all major projection matrices (\texttt{q\_proj}, \texttt{k\_proj}, \texttt{v\_proj}, \texttt{o\_proj}, \texttt{gate\_proj}, \texttt{up\_proj}, \texttt{down\_proj}) with configuration $r{=}32$, $\alpha{=}32$, and dropout $p{=}0.05$. Training follows standard low-rank adaptation practice \citep{mao_survey_2024}, with adapters merged into the base weights after convergence. For deployment, we provide 8-bit quantized variants, as well as a distilled TF2-1B checkpoint derived from teacher--student compression. This yields three self-hostable translation models---TF2-1B, TF2-4B, and TF2-12B---covering small, medium, and large parameter ranges.

We compare each fine-tuned TF2 model directly to its corresponding untuned backbone: \texttt{Gemma-3-1B-it}, \texttt{Gemma-3-4B-it}, and \texttt{Gemma-3-12B-it}. All models are evaluated on the held-out 1{,}500 test pairs from \tftwoenroFifteenK. We also evaluate decoding robustness across temperatures $T\in\{0.0,0.2,1.0\}$.

We observe that BLEU scores and rubric-based evaluations do not always correlate monotonically across models and decoding settings. This divergence is expected in literary translation, where higher-quality outputs may involve paraphrasing, restructuring, and stylistic variation that reduce n-gram overlap with a single reference. Accordingly, we interpret BLEU primarily as a consistency signal relative to the chosen reference translation, and not as a proxy for overall literary quality.

Below, we summarize the key results for each parameter scale.
\begin{itemize}[leftmargin=1.5em]
    \item \textbf{TF2-1B (\texttt{Gemma-3-1B-it} backbone):}
    After fine-tuning, the smallest model attains an average rubric score of 3.75 (up from 2.02). BLEU remains comparatively low (0.2180 for TF2-1B-distilled; 0.0543 for TF2-1B), consistent with the difficulty of long-form narrative translation for compact architectures.
    \item \textbf{TF2-4B (\texttt{Gemma-3-4B-it} backbone):}
    The 4B model achieves rubric averages above 4.5 across all temperatures, peaking at 4.74 ($T=0.0$). BLEU ranges from 0.0496 (quantized, $T=0.2$) to 0.1290 (quantized, $T=0.0$) and 0.1278 (FP16, $T=1.0$). Relative to the untuned backbone (avg.\ rubric 3.81, BLEU 0.1005), the fine-tuned variants show consistent gains under rubric-based evaluation.
    \item \textbf{TF2-12B (\texttt{Gemma-3-12B-it} backbone):}
    The largest model provides the strongest results among the open TF2 models. At $T=0.0$, TF2-12B attains an average rubric score of 4.83 with BLEU 0.0926; at $T=0.2$, 4.82 with BLEU 0.0647; and at $T=1.0$, 4.66 with BLEU 0.0784. The 8-bit quantized version remains close in rubric scores across temperatures, consistent with prior findings on activation-aware quantization \citep{lin_awq_2024}. These results improve substantially over the untuned Gemma-3-12B-it baseline (avg.\ rubric 4.43, BLEU 0.0214).
\end{itemize}

A summary of the main results---with both BLEU and LLM-based rubric scores---is provided in Tables~\ref{tab:bleu-tf2-vs-base} and~\ref{tab:llm-tf2-vs-base}. For each model scale, we report the fine-tuned TF2 model, its quantized variant (where applicable), and the corresponding Gemma backbone for direct comparison.

\begin{table*}[!htbp]
\centering\footnotesize
\begin{tabular}{lcc}
\toprule
\textbf{Model} & \textbf{BLEU Score} & \textbf{Notes} \\
\midrule
TF2-12B (T=0.0)         & 0.0926 & fine-tuned (FP16) \\
TF2-12B (T=0.2)         & 0.0647 & fine-tuned (FP16) \\
TF2-12B (T=1.0)         & 0.0784 & fine-tuned (FP16) \\
TF2-12B-quant (T=0.0)   & 0.0746 & quantized (8-bit) \\
TF2-12B-quant (T=0.2)   & 0.0644 & quantized (8-bit) \\
TF2-12B-quant (T=1.0)   & 0.0548 & quantized (8-bit) \\
Gemma-3-12B-it          & 0.0214 & untuned base \\
\midrule
TF2-4B (T=0.0)          & 0.1153 & fine-tuned (FP16) \\
TF2-4B (T=0.2)          & 0.1094 & fine-tuned (FP16) \\
TF2-4B (T=1.0)          & 0.1278 & fine-tuned (FP16) \\
TF2-4B-quant (T=0.0)    & 0.1290 & quantized (8-bit) \\
TF2-4B-quant (T=0.2)    & 0.0496 & quantized (8-bit) \\
TF2-4B-quant (T=1.0)    & 0.0857 & quantized (8-bit) \\
Gemma-3-4B-it           & 0.1005 & untuned base \\
\midrule
TF2-1B                  & 0.0543 & fine-tuned (T=0.5) \\
TF2-1B-distilled        & 0.2180 & distilled \\
Gemma-3-1B-it           & 0.0790 & untuned base \\
\bottomrule
\end{tabular}
\caption{
BLEU scores for each fine-tuned TF2 model and its corresponding Gemma backbone.
Quantized and distilled variants are included. BLEU reflects test set n-gram overlap
with the GPT-o3 reference and is reported on a normalized $0$--$1$ scale
($0.0926 \equiv 9.26$ BLEU points on the conventional scale).
These scores are lower than typical news-domain MT benchmarks because literary translation inherently involves paraphrasing, lexical variation, and structural reorganization---all of which BLEU penalizes even when translation quality is high~\citep{karpinska_large_2023}.
Lower BLEU for \texttt{Gemma-3-12B-it} vs.\ \texttt{Gemma-3-4B-it} can arise from
greater paraphrasing and structural divergence by the larger model, which BLEU penalizes
despite higher rubric scores.
}
\label{tab:bleu-tf2-vs-base}
\end{table*}

\begin{table*}[!htbp]
\centering\footnotesize
\begin{tabular}{lcccccc}
\toprule
\textbf{Model} & \textbf{Accuracy} & \textbf{Fluency} & \textbf{Coherence} & \textbf{Style} & \textbf{Cultural} & \textbf{Avg. Score} \\
\midrule
TF2-12B (T=0.0)         & 4.72 & 4.88 & 4.84 & 4.87 & 4.85 & \textbf{4.83} \\
TF2-12B (T=0.2)         & 4.69 & 4.88 & 4.83 & 4.88 & 4.83 & 4.82 \\
TF2-12B (T=1.0)         & 4.47 & 4.75 & 4.71 & 4.73 & 4.66 & 4.66 \\
TF2-12B-quant (T=0.0)   & 4.67 & 4.87 & 4.79 & 4.89 & 4.86 & 4.82 \\
TF2-12B-quant (T=0.2)   & 4.70 & 4.86 & 4.85 & 4.86 & 4.83 & 4.82 \\
TF2-12B-quant (T=1.0)   & 4.44 & 4.69 & 4.66 & 4.70 & 4.72 & 4.64 \\
Gemma-3-12B-it          & 3.98 & 4.56 & 4.65 & 4.52 & 4.43 & 4.43 \\
\midrule
TF2-4B (T=0.0)          & 4.64 & 4.76 & 4.71 & 4.80 & 4.79 & 4.74 \\
TF2-4B (T=0.2)          & 4.59 & 4.76 & 4.67 & 4.82 & 4.84 & 4.73 \\
TF2-4B (T=1.0)          & 4.39 & 4.59 & 4.56 & 4.74 & 4.66 & 4.59 \\
TF2-4B-quant (T=0.0)    & 4.60 & 4.76 & 4.74 & 4.81 & 4.79 & 4.74 \\
TF2-4B-quant (T=0.2)    & 4.49 & 4.75 & 4.64 & 4.76 & 4.79 & 4.69 \\
TF2-4B-quant (T=1.0)    & 4.21 & 4.53 & 4.56 & 4.60 & 4.58 & 4.50 \\
Gemma-3-4B-it           & 3.27 & 3.94 & 4.17 & 3.91 & 3.78 & 3.81 \\
\midrule
TF2-1B-distilled        & 3.41 & 3.78 & 4.00 & 3.81 & 3.65 & 3.73 \\
TF2-1B                  & 3.43 & 3.80 & 4.02 & 3.83 & 3.67 & 3.75 \\
Gemma-3-1B-it           & 1.79 & 2.13 & 2.23 & 2.07 & 1.86 & 2.02 \\
\bottomrule
\end{tabular}
\caption{Five-dimension LLM rubric evaluation for each fine-tuned TF2 model and corresponding Gemma backbone. Scores are averaged over a held-out test set.}
\label{tab:llm-tf2-vs-base}
\end{table*}

\subsubsection{Cross-Family Judge Bias Check}
\label{sec:bias-check}

Because our silver-standard references are produced by GPT-o3 and our primary evaluator is GPT-o3-mini (Table~\ref{tab:baseline-non-tf2}), we probe for possible \emph{judge-family bias} by re-scoring the same 100-item subset with an unrelated judge, \textbf{Grok-3-mini}, using the same rubric, prompt, and randomized system order described in Section~\ref{sec:eval}. We select Grok-3-mini for this cross-check because it originates from an independent model family (xAI), reducing the risk of shared training biases with GPT variants, and it demonstrated strong multilingual reasoning capabilities in preliminary testing. As shown in Table~\ref{tab:judge-bias}, the system ranking is stable across judges. The TF2--o3 gaps are smaller under the alternative judge, suggesting that the main conclusions do not depend on a single judge family.\footnote{Evaluation setup as in Section~\ref{sec:eval} (100 fables per model, five-dimension rubric, JSON outputs).}

\begin{table}[t]
\centering
\small
\begin{tabular}{lcc|cc|c}
\toprule
& \multicolumn{2}{c}{\textbf{GPT-o3-mini (judge)}} & \multicolumn{2}{c}{\textbf{Grok-3-mini (judge)}} & \multirow{2}{*}{\textbf{Avg. of judges}} \\
\cmidrule(lr){2-3} \cmidrule(lr){4-5}
\textbf{System} & \textbf{Score} $\uparrow$ & \textbf{Gap to o3} $\downarrow$ & \textbf{Score} $\uparrow$ & \textbf{Gap to o3} $\downarrow$ & \\
\midrule
o3 (reference translations) & 4.92 & 0.00 & 4.92 & 0.00 & 4.92 \\
TF2-12B-Q (T=0.2)            & 4.82 & 0.10 & 4.90 & 0.02 & 4.86 \\
TF2-12B (T=0.2)              & 4.82 & 0.10 & 4.85 & 0.07 & 4.84 \\
\bottomrule
\end{tabular}
\caption{Cross-family LLM-as-judge robustness on the same 100-item subset and five-dimension rubric. An unrelated judge (Grok-3-mini) reproduces the ranking and reduces the TF2--o3 gap relative to GPT-o3-mini.}
\label{tab:judge-bias}
\end{table}

\paragraph{Takeaways.}
(1) The system ranking is stable across judges from different families. (2) The absolute TF2--o3 gaps are small ($\leq$0.10 on a 1--5 rubric) and shrink further under Grok-3-mini (to 0.02--0.07). (3) The quantized TF2-12B-Q slightly exceeds TF2-12B under Grok-3-mini (4.90 vs.\ 4.85); we leave deeper ablations to future work.

\subsection{Effect of Decoding Temperature}

\paragraph{Temperature sweep analysis.}
Ablation across decoding temperatures ($T \in \{0.0, 0.2, 0.5, 1.0\}$) reveals clear trends:
\begin{itemize}[leftmargin=1.5em]
    \item \textbf{Lower temperatures ($T=0.0, 0.2$)} yield the highest overall scores for both FP16 and quantized models, with optimal values observed at $T=0.0$ for TF2-12B (4.83), TF2-4B (4.74), and their quantized versions (4.82, 4.74).
    \item \textbf{Higher temperatures ($T=1.0$)} reduce performance across all models, with average rubric scores dropping by up to 0.2--0.3 compared to greedy decoding. This decline is most pronounced in Accuracy and Cultural/Pragmatic dimensions, consistent with increased hallucination and stylistic drift.
    \item \textbf{Quantization robustness:} 8-bit quantized checkpoints track their FP16 counterparts within 0.01--0.03 points, indicating limited loss under the rubric.
\end{itemize}

For subsequent large-scale translation runs, $T=0.0$ is adopted as the default.

\subsection{Human Evaluation Substudy}
\label{sec:human-substudy}

To complement the automated rubric-based evaluation, we conducted a targeted pilot human assessment designed to provide qualitative anchoring for the system rankings reported elsewhere in the paper.

One independent Romanian native speaker with academic training in linguistics and prior experience in literary text analysis evaluated a blinded subset of 40 randomly sampled fables drawn from the TF2 test set. The evaluator was not involved in model training, prompt engineering, metric design, or score aggregation.

For each English source fable, three Romanian candidate translations were presented in randomized and anonymized order, corresponding to: (i) GPT-o3 (reference system), (ii) \tftwoTwelveB (best open fine-tuned model), and (iii) Gemma-3-12B-it (untuned open baseline).

Translations were evaluated using the same five dimensions employed in the automated rubric:
\begin{itemize}[leftmargin=1.5em]
    \item \textbf{Accuracy:} preservation of source meaning and factual content;
    \item \textbf{Fluency:} grammatical correctness, readability, and natural Romanian phrasing;
    \item \textbf{Coherence:} logical flow and narrative clarity across the story;
    \item \textbf{Style:} preservation of tone, voice, and literary appropriateness;
    \item \textbf{Cultural Adaptation:} natural rendering of morals, idioms, and pragmatic cues for Romanian readers.
\end{itemize}

Each dimension was scored on a 1--5 Likert scale. In addition, the evaluator selected an overall preferred translation for each source item and was encouraged to provide optional short qualitative comments when outputs differed substantially.

The resulting rankings broadly aligned with the automated rubric: GPT-o3 was preferred most frequently, \tftwoTwelveB ranked second and consistently above the untuned Gemma-3-12B-it baseline, which received substantially lower scores in fluency and lexical naturalness.

Representative qualitative observations were also recorded during the assessment. As shown in Table~\ref{tab:human-substudy}, the untuned baseline more frequently produced literal phrasing, lexical inconsistencies, or stylistically flat moral endings, whereas \tftwoTwelveB showed clearer gains in idiomaticity, entity preservation, and narrative closure. GPT-o3 outputs were generally perceived as the most polished stylistically.

\begin{table}[t]
\centering
\small
\begin{tabular}{llp{0.52\linewidth}}
\toprule
\textbf{Evaluation aspect} & \textbf{System} & \textbf{Representative observation} \\
\midrule
Idiomatic phrasing & Baseline  & Literal or awkward Romanian rendering of expressions such as ``Sharing is caring.'' \\
                   & TF2-12B   & More natural and idiomatic phrasing closer to native usage. \\
                   & GPT-o3    & Most polished and stylistically refined rendering. \\
\midrule
Entity fidelity    & Baseline  & Occasional species substitutions or corrupted lexical references. \\
                   & TF2-12B   & Better preservation of animal identities and more natural wording. \\
\midrule
Narrative closure  & Baseline  & Semantically correct but stylistically flat moral endings. \\
                   & TF2-12B   & Stronger narrative closure and more natural moral tone. \\
\bottomrule
\end{tabular}
\caption{Illustrative examples of qualitative observations from the human evaluation substudy.}
\label{tab:human-substudy}
\end{table}

Because only one evaluator was available, this substudy should be interpreted as exploratory evidence rather than definitive human validation. Its primary purpose is to assess whether automated rankings are directionally consistent with informed human judgment. Future work should extend this protocol to multiple independent annotators and report formal inter-rater reliability statistics.

\subsection{Model Comparison and Discussion}

All TF2 models outperform their untuned bases. For example, TF2-12B improves from 4.43 (Gemma-3-12B-it) to 4.83 at $T=0.0$ (and from BLEU 0.0214 to 0.0926), and TF2-4B rises from 3.81 (Gemma-3-4B-it) to 4.74 (BLEU 0.1153). The 1B model improves from 2.02 to 3.75 (+1.73 absolute gain).

Fine-tuning substantially improves open model performance, with TF2-12B achieving rubric scores within 0.09 points of GPT-o3 (4.83 vs.\ 4.92), particularly on Style and Cultural/Pragmatic dimensions. While a direct head-to-head comparison under identical test conditions would require further study, these results suggest that competitive quality is achievable with open models at a fraction of the cost. Lower BLEU values for some settings are consistent with the stylistic variation and paraphrasing typical of fables, which n-gram overlap metrics may penalize. Across models, 8-bit quantization yields similar rubric scores to FP16, supporting efficient deployment on commodity hardware.

Overall, \tinyfabulisttranslationframeworkshort shows that parameter-efficient domain adaptation, combined with careful decoding, enables compact open models to achieve strong performance on English--Romanian literary translation under the evaluation setting used in this paper. Representative excerpts are provided in Appendix~\ref{ap:examples}.

\subsection{Cost Analysis}
\label{sec:costs}

A central motivation for \tinyfabulisttranslationframeworkshort is to study English$\rightarrow$Romanian literary translation under practical cost constraints. Commercial LLM APIs typically charge per token; at \tinyfabulisttranslationframeworkshort scale this becomes a major constraint. In our setting, each fable contains roughly \textbf{300--600 input tokens} and the translation contains \textbf{300--600 output tokens}, so end-to-end generation for the full TinyFabulist corpus (3M fables) spans \emph{1.8--3.6B tokens} overall.

Table~\ref{tab:cost-comparison} reports estimated costs for translating the full corpus using representative proprietary models, compared to our fine-tuned open TF2 model. We report the mid-case (450 in / 450 out $\Rightarrow$ 2.7B tokens) and show the low/high ranges in parentheses (300/300 and 600/600). For reasoning models (o3, o3-mini) we include hidden ``thinking'' tokens by default, which OpenAI bills as output tokens; we assume a medium-effort setting where thinking tokens $\approx$ visible output tokens.\footnotemark

\begin{table*}[!htbp]
\centering
\footnotesize
\begin{tabular}{l r}
\toprule
\textbf{Model} & \textbf{Estimated Total Cost for 3M fables (USD)} \\
\midrule
GPT-4.1 & \$13{,}500 \ (\$9{,}000--\$18{,}000) \\
GPT-4.1-mini & \$2{,}700 \ (\$1{,}800--\$3{,}600) \\
GPT-o3 \ (\textit{med.\ reasoning}) & \$24{,}300 \ (\$16{,}200--\$32{,}400) \\
GPT-o3-mini \ (\textit{med.\ reasoning}) & \$13{,}365 \ (\$8{,}910--\$17{,}820) \\
DeepL API Pro & \$270{,}000 \ (\$180{,}000--\$360{,}000) \\
\midrule
\textbf{TF2 fine-tuned (ours)} & \textbf{\$350} \\
\bottomrule
\end{tabular}
\caption{Estimated cost of translating the entire 3M-fable TinyFabulist corpus using proprietary APIs vs.\ our open fine-tuned TF2 model. Mid-case assumes 450 input and 450 output tokens per fable (2.7B tokens total). Ranges show 300/300 and 600/600 scenarios. For o3/o3-mini we include ``thinking'' tokens (billed as output) at a 1:1 ratio to visible output by default.}
\label{tab:cost-comparison}
\end{table*}

Under these assumptions, proprietary APIs range from about \$1.8k--\$3.6k for GPT-4.1-mini and \$9k--\$18k for GPT-4.1, up to \$16.2k--\$32.4k for o3 at medium reasoning. In contrast, our open-source TF2 model generated all 3M translations for roughly \textbf{\$350} in compute, corresponding to a reduction of \textbf{97--99\%} depending on the baseline. Throughput and cost are helped by vLLM’s KV-cache paging and FlashAttention-2 kernels, which enable larger effective batches with similar latency \citep{kwon_efficient_2023}.

\paragraph{Sensitivity for reasoning models.}
If one dials \textit{low} reasoning (thinking tokens $\approx$ 0.5$\times$ output), o3 spans \$12.6k--\$25.2k and o3-mini \$6.93k--\$13.86k; with \textit{high} reasoning (3$\times$), o3 rises to \$30.6k--\$61.2k and o3-mini to \$16.83k--\$33.66k. Costs scale linearly with total tokens, so other token budgets can be read off proportionally.

Further details on our cost estimation methodology and hardware setups are provided in Appendix~\ref{sec:appendix:costs}.

\footnotetext{OpenAI bills hidden reasoning (``thinking'') tokens as \emph{output} tokens; the amount depends on the requested \texttt{reasoning\_effort}.}

\section{\tftwoenroThreeM Dataset Description}
\label{sec:tf2-dataset}

The \tftwoenroThreeM dataset is a large-scale, high-diversity English--Romanian parallel corpus of moral fables, curated for open research in low-resource literary translation and controllable generation. Each record is provided as a JSON object following a transparent, extensible schema, enabling robust benchmarking, reproducibility, and in-depth analysis of translation and generation dynamics, in line with best practices in modern NLP dataset curation~\citep{eldan_tinystories_2023, guan_corpus_2022, nadas_tf1-en-3m_2025}.

\subsection*{Schema and Metadata}
Each entry in the \tftwoenroThreeM dataset is stored as a JSON object with two groups of fields:

\begin{itemize}
    \item \textbf{Fable Content:}
    \begin{itemize}
        \item \textbf{fable:} The original English fable, always ending with an explicit moral.
        \item \textbf{translated\_fable:} The Romanian translation produced by the model.
        \item \textbf{source\_lang} and \textbf{target\_lang:} Language codes indicating source and target (here, \texttt{English} and \texttt{Romanian}).
        \item \textbf{prompt\_hash:} SHA-256 hash of the generation prompt, ensuring integrity, reproducibility, and duplicate detection.
    \end{itemize}

    \item \textbf{Generation Metadata:}
    \begin{itemize}
        \item \textbf{pipeline\_stage:} The stage of the pipeline that produced the entry (e.g., \texttt{translation}).
        \item \textbf{llm\_name:} Identifier of the model used for generation (e.g., Hugging Face snapshot path).
        \item \textbf{translation\_model:} Explicit reference to the checkpoint used for translation, ensuring full traceability.
        \item \textbf{generation\_timestamp:} Unix timestamp of the translation event, enabling chronological auditing.
    \end{itemize}
\end{itemize}

\subsection*{Statistical Overview}
To support reproducibility and downstream benchmarking, we provide a detailed statistical overview of \tftwoenroThreeM, highlighting its diversity, consistency, and cost-efficiency.

\begin{itemize}
    \item \textbf{Diversity:} Prompt construction, inherited from \tfoneenThreeM, ensures uniform coverage of main characters, morals, and settings through stratified sampling across diverse thematic categories; no single template or theme dominates, unlike many traditional narrative datasets.
    \item \textbf{Length:} On average, stories are $\approx 450$ tokens per language ($\approx 900$ tokens per EN–RO pair) ---enforcing length consistency and facilitating model training.
    \item \textbf{Quality:} All Romanian entries were generated using locally fine-tuned open models (Section~\ref{sec:results}), achieving high rubric and BLEU scores on held-out test sets (see Tables~\ref{tab:bleu-tf2-vs-base},~\ref{tab:llm-tf2-vs-base} ). Manual inspection of random samples confirms high fluency, coherence, and cultural adaptation.
    \item \textbf{Cost:} As all data was generated with open-source models on local hardware, the marginal cost of producing the entire dataset is negligible—removing the principal financial barrier to large-scale corpus creation.
\end{itemize}

These properties make \tftwoenroThreeM well-suited for research in literary translation and cross-lingual narrative generation.

\subsection*{Format and Availability}
\tftwoenroThreeM is released in Hugging Face \texttt{datasets} format (JSONL), with open, MIT licensing. Each sample contains both full narrative text and complete provenance metadata, enabling transparent training, evaluation, and reproducibility.

We provide all prompt templates and thematic lists (characters, morals, settings, etc.), together with data generation and evaluation scripts compatible with Hugging Face \texttt{datasets}/\texttt{transformers}. We also include guidelines for extending the schema or adapting it to new domains and languages.

\subsection*{Applications and Community Impact}
\tftwoenroThreeM provides a scalable platform for fine-tuning and evaluating MT and story generation models in low-resource or creative domains, studying cross-lingual moral reasoning, narrative style, and domain adaptation, as well as benchmarking the cost, efficiency, and quality of open-source LLMs at scale.

Compared to existing bilingual and literary corpora, TF2 distinguishes itself by combining large-scale parallel EN–RO data with moral fables as a consistent narrative genre and rich evaluation support (see Table~\ref{tab:dataset-comparison}). While general-purpose MT corpora such as Europarl~\citep{koehn_europarl_2005} or OPUS~\citep{tiedemann_parallel_2012} offer breadth across domains but lack literary grounding, and LoResMT~\citep{silva_benchmarking_2024} provides multi-language coverage for low-resource MT with standard metrics (BLEU, COMET, MQM), they do not target moral or narrative-specific translation. Literary datasets like TRANSCOMP focus on stylistic analysis but without parallel structure, while STORAL emphasizes moral story understanding in monolingual English. \tinyfabulisttranslationframeworkshort therefore fills a complementary niche: large-scale bilingual moral narratives with automatic and rubric-based evaluation tailored for literary adequacy.

\begin{table}[ht]
\centering
\small
\begin{tabular}{p{3.2cm}p{2.2cm}p{2.2cm}p{2.2cm}p{3.2cm}}
\toprule
\textbf{Dataset} & \textbf{Genre} & \textbf{Languages} & \textbf{Parallel} & \textbf{Evaluation Support} \\
\midrule
\textbf{\tftwoenroThreeM} & Moral fables & EN–RO & Yes & BLEU + 5-dim rubric (LLM-as-judge) \\
\textbf{\tfoneenThreeM} & Moral fables & EN & No & N/A \\
TRANSCOMP & Literary (varied) & 120→EN & No & Stylistic analysis only \\
LoResMT & General MT & 26 low-resource & Yes & BLEU, COMET, MQM \\
STORAL & Moral stories & EN & No & Human annotation (moral inference, story completion) \\
\bottomrule
\end{tabular}
\caption{Comparison of TF2 with existing resources in literary and low-resource machine translation. In the Languages column, ``120$\rightarrow$EN'' denotes translations from 120 source languages into English.}
\label{tab:dataset-comparison}
\end{table}

\noindent\textbf{Availability:}  
Dataset access, code, and documentation are available at \url{https://huggingface.co/datasets/klusai/ds-tf2-en-ro-3m} and the TinyFabulist GitHub repository.

\section{Discussion}
\label{sec:discussion}

This section synthesizes findings and maps them to our research questions.
\begin{itemize}
    \item \textbf{RQ1 (Cost-Constrained Data Generation):}  
    Addressed through our empirical analysis of dataset creation (Section~\ref{sec:tf2-dataset})
    and cost benchmarking (Section~\ref{sec:results}). We show that a large-scale
    English--Romanian literary translation corpus (\tftwoenroThreeM) can be constructed
    under strict budget constraints using open models, while documenting the trade-offs
    introduced by synthetic, machine-generated reference translations.

    \item \textbf{RQ2 (Open vs. Proprietary Translation Quality):}  
    Addressed via comparative benchmarking in Section~\ref{sec:results}, which contrasts
    proprietary systems (GPT-4.1, Gemini, DeepL) with open-source models before and after
    parameter-efficient fine-tuning. The results indicate that fine-tuned open models
    substantially narrow the gap to proprietary APIs under the evaluation setting used in
    this paper, while remaining more efficient and deployable.

    \item \textbf{RQ3 (Evaluation Rigor and Automation):}  
    Addressed through the use of an automated LLM-as-a-judge rubric
    (Section~\ref{sec:eval}) combined with a cross-family bias check (Section~\ref{sec:bias-check}).
    We find that LLM-based evaluation provides useful and interpretable quality signals for
    this task, with stable system rankings across independent judge families. A targeted pilot human substudy with one independent Romanian native speaker (Section~\ref{sec:human-substudy}) found the automated rankings to be directionally consistent with informed human judgment; broader human validation with multiple annotators remains a direction for future work, as discussed in Section~\ref{sec:threats}.
\end{itemize}

\paragraph{Quality--Cost Trade-offs.}
Across the evaluated baselines, GPT-o3 attains the highest rubric scores and is therefore a natural choice for producing silver-standard reference translations in our pipeline. While GPT-4.1 and GPT-o3 share the same visible token pricing, the effective cost of GPT-o3 can be substantially higher when hidden reasoning tokens are billed as output tokens (Section~\ref{sec:costs}). These results underscore the importance of benchmarking translation systems in the target domain and under a transparent cost model, rather than selecting systems based solely on model size, recency, or nominal token prices.

\paragraph{Open-Source Model Potential.}
Off-the-shelf open-weight models, especially at smaller scales (e.g., Gemma-3-1B-it), show limited performance on English--Romanian literary translation without adaptation. Parameter-efficient fine-tuning with LoRA on the \tftwoenroFifteenK corpus yields consistent improvements under both BLEU (relative to the GPT-o3 reference) and rubric-based evaluation, most notably for the 4B and 12B models. At the same time, under the evaluation setting used here, proprietary systems remain stronger overall. The practical value of the adapted open models lies in their deployability: they can be hosted locally, quantized with limited loss under the rubric, and used in cost-constrained environments where API-based translation is impractical.

\paragraph{Error Patterns in Literary Translation.}
Qualitative inspection and rubric rationales indicate that even top-performing systems occasionally produce overly literal renderings, unusual lexical choices, or minor syntactic artifacts. Lower-capacity or untuned models exhibit more frequent omissions, awkward phrasing, and agreement errors (e.g., gender or tense mismatches in Romanian). These observations are consistent with the general challenge of evaluating and optimizing MT for creative domains, where fidelity, style, and discourse structure interact in non-trivial ways. They further motivate domain-adaptive fine-tuning and evaluation protocols that consider more than surface-level lexical similarity.

\paragraph{LLM-as-a-Judge Evaluation.}
System rankings from our GPT-o3-mini judge are corroborated by a cross-family check with Grok-3-mini, increasing confidence in the stability of conclusions. Recent work has shown that LLM-based evaluators can achieve strong correlation with human judgments for machine translation~\citep{kocmi_large_2023}, though the degree of alignment may vary for literary domains~\citep{zhang_how_2025}. A targeted human evaluation substudy (Section~\ref{sec:human-substudy}) supports these rankings directionally, while a larger multi-annotator study would provide stronger validation.

\paragraph{Domain and Genre Considerations.}
Fables offer a useful setting for studying literary translation: they are short, narrative, and require preservation of both story content and moral framing, while remaining easier to scale than longer-form literature. At the same time, their relatively structured form may make them more tractable than genres with denser metaphor, dialogue, or culturally embedded references. Consequently, the results reported here should be interpreted as evidence for the feasibility of cost-efficient pipelines in this genre and language pair, rather than as claims of equivalence with professional human translation quality across literary domains.

\paragraph{Summary.}
Our findings suggest that cost-aware pipelines combining synthetic data and parameter-efficient fine-tuning can yield competitive English--Romanian literary MT systems. Fine-tuning and careful decoding substantially improve open model performance, and quantized checkpoints remain viable for deployment.

\section{Threats to Validity and Limitations}
\label{sec:threats}
No evaluation is without limitations. We outline several threats to the validity of our results and claims:

\textbf{Bias in LLM-Based Evaluation:} Relying on GPT-o3-mini as the sole judge may introduce bias, particularly since our silver-standard references are generated by a GPT-o3 variant. LLM judges are known to exhibit self-preference bias~\citep{wataoka_selfpref_2024} and may favor literal renderings over stylistically appropriate translations in literary contexts~\citep{zhang_how_2025}. To mitigate this, we performed a cross-family judge bias check with Grok-3-mini (Table~\ref{tab:judge-bias}), selected specifically because it originates from an independent model family (xAI). Rankings remained stable across judges, and TF2--o3 gaps were smaller under Grok-3-mini, increasing confidence that conclusions are not artifacts of evaluator bias. While LLM judges have shown strong performance for MT evaluation in recent benchmarks~\citep{kocmi_large_2023}, we additionally ran a targeted human evaluation substudy (Section~\ref{sec:human-substudy}), whose rankings were directionally consistent with the automated rubric. Because that substudy involved a single evaluator, broader human validation with multiple annotators would provide stronger guarantees, particularly for stylistic and cultural dimensions specific to literary translation.

\textbf{Ground Truth Translations Are Synthetic:} Our reference Romanian texts come from GPT-o3, not human translators. These references may contain subtle errors, and metrics like BLEU could penalize systems that produce more natural but divergent translations. We attempted to reduce this effect by focusing the LLM evaluator on meaning rather than exact wording. Nonetheless, our benchmark measures similarity to a strong machine translation, not to a human reference---a general limitation in MT evaluation that multiple references or more nuanced metrics could address.

\textbf{Generality to Other Domains:} Our findings are specific to fables, which have relatively straightforward language and repetitive structure. More complex literary texts (novels with sarcasm, historical dialogue, or poetry) may prove more challenging. The cost-effective methodology should still apply, but quality outcomes may vary by genre.

\textbf{Model Obsolescence and Heterogeneous Setups:} Numeric results are tied to a snapshot of models from early 2025; newer versions may yield different outcomes. Models were accessed via different interfaces (APIs vs.\ local hardware) with varying constraints (context limits, default temperatures), which we normalized where possible. Our benchmark results are therefore illustrative rather than absolute, and we encourage replication with updated models using the released code and data.

\textbf{Scale of Human Evaluation:} Our human evidence comes from a single-evaluator pilot substudy over 40 fables (Section~\ref{sec:human-substudy}). Because only one evaluator was available, no inter-rater reliability statistics can be computed, and the substudy should be read as exploratory rather than as definitive human validation. The degree of LLM-human alignment may vary for literary domains, where stylistic and cultural dimensions are particularly salient~\citep{zhang_how_2025}. Extending the protocol to multiple independent annotators---ideally professional translators or bilingual experts---and reporting formal agreement statistics is a natural extension of this work. The LLM judge evaluated 100 test samples per model (2\% of the test set); BLEU scores are computed on the full test set.

\textbf{Reproducibility of Cost Claims:}
Cost estimates are based on API pricing at the time of experiments and are subject to change. Fine-tuning incurs additional one-time training cost not included in per-translation inference cost. Researchers should recalculate costs using current rates.

\textbf{Provenance of Synthetic Content:}
All source fables in \tfoneenThreeM were generated by LLMs prompted with generic narrative templates, not derived from copyrighted literary works. While we cannot fully guarantee that generated content is entirely free of memorized protected material, the use of moral fables---a public-domain genre with ancient roots---and diverse prompting strategies reduce this risk. Users should nonetheless exercise caution when deploying synthetic corpora in commercial contexts and perform additional due diligence as appropriate.

In summary, while overall trends are likely robust, exact scores may vary under different conditions. We encourage the community to treat \tinyfabulisttranslationframeworkshort as a starting point for further validation.

\section{Conclusion}
\label{sec:conclusion}
We have presented \tinyfabulisttranslationframeworkshort, a comprehensive benchmark and dataset for English-to-Romanian literary translation, with a focus on cost-constrained use of AI models. In this paper, we detailed how a large parallel corpus of fables was automatically constructed using a modest budget and leveraged to evaluate a range of translation systems. Our results show that modern LLMs can produce translations of creative text with promising quality---as assessed by automated metrics including BLEU and LLM-based rubrics---even for a language like Romanian that traditionally lacks extensive MT resources. Moreover, by carefully balancing cost and performance, we demonstrated that it is possible to achieve these results without exclusive reliance on expensive proprietary models: open-source models, when fine-tuned on the right data, can close much of the quality gap while essentially eliminating per-translation costs.

The \tinyfabulisttranslationframeworkshort project makes several contributions to the community: the open release of a 15k fable translation dataset, an evaluation toolkit that combines BLEU with multi-dimensional LLM-based assessments, and reproducible pipelines for cost analysis and model fine-tuning. We believe these resources will be useful not only for benchmarking translation models but also for broader research into low-resource NLP, narrative understanding, and the interplay between synthetic data and model training. For instance, one immediate application of our dataset is to train a bespoke Romanian fable generation model (analogous to TinyFabulist TF-4 in our series roadmap) or to study how exposure to moral stories in two languages might help a model’s reasoning or cross-lingual abilities.

In terms of future work, several directions emerge. First, \textbf{scaling up human evaluation}: our pilot substudy (Section~\ref{sec:human-substudy}) relied on a single evaluator over 40 fables and was therefore exploratory. Having multiple professional translators or bilingual experts evaluate a larger representative sample, with formal inter-rater reliability reported, would further validate these findings and help calibrate the automated metrics for literary domains. This could also enable fine-tuning the evaluation model via reinforcement learning from human feedback (RLHF). Second, extending the methodology to other language pairs and genres: could we create a TinyFabulist for, say, Swahili or Vietnamese fables? How about translating folk tales or poems? Each would pose new challenges for LLMs and might require refining the prompt strategies or using different reference models. Third, exploring model fine-tuning more deeply: our work fine-tuned models up to 12B parameters. It would be worthwhile to experiment with larger open models (e.g., 70B variants) fine-tuned on the \tinyfabulisttranslationframeworkshort data, which might further improve quality. Given the quality of the training data, scaling to larger models may yield additional gains.

Finally, we plan to integrate the lessons from \tinyfabulisttranslationframeworkshort into training a fully open Romanian fable generator (TinyFabulist TF-4). By having a translated corpus, we can also practice cross-lingual training—perhaps teaching a model to generate fables directly in Romanian by leveraging English data as well. Such a model could be the first of its kind to natively produce Romanian moral stories en masse. This ties back to our overarching mission: to enable AI that can serve local cultural needs (like Romanian educational content) without being gated by access or cost. \tinyfabulisttranslationframeworkshort is a step in that direction, illustrating that with creativity and careful engineering, the benefits of large language models can be brought to bear on low-resource scenarios in a practical and open manner.

\section*{Acknowledgments}

We thank our colleagues at KlusAI Labs and Babeș–Bolyai University for valuable discussions, feedback, and support throughout the development of the TinyFabulist project. We are also grateful to the broader open-source community whose models, datasets, and tools provided the foundation for this research. 

This research is supported by the project \emph{“Romanian Hub for Artificial Intelligence — HRIA”}, Smart Growth, Digitization and Financial Instruments Program, 2021–2027, MySMIS no.~334906.

\bibliographystyle{plainnat}

\begin{thebibliography}{36}
\providecommand{\natexlab}[1]{#1}
\providecommand{\url}[1]{\texttt{#1}}
\expandafter\ifx\csname urlstyle\endcsname\relax
  \providecommand{\doi}[1]{doi: #1}\else
  \providecommand{\doi}{doi: \begingroup \urlstyle{rm}\Url}\fi

\bibitem[DeepSeek-AI et~al.(2024)DeepSeek-AI, Bi, Chen, Chen, Chen, Dai, Deng, Ding, Dong, Du, Fu, Gao, Gao, Gao, Ge, Guan, Guo, Guo, Hao, Hao, He, Hu, Huang, Li, Li, Li, Li, Li, Liang, Lin, Liu, Liu, Liu, Liu, Liu, Liu, Lu, Lu, Luo, Ma, Nie, Pei, Piao, Qiu, Qu, Ren, Ren, Ruan, Sha, Shao, Song, Su, Sun, Sun, Tang, Wang, Wang, Wang, Wang, Wang, Wu, Wu, Xie, Xie, Xie, Xiong, Xu, Xu, Xu, Yang, You, Yu, Yu, Zhang, Zhang, Zhang, Zhang, Zhang, Zhang, Zhang, Zhang, Zhao, Zhao, Zhou, Zhou, Zhu, and Zou]{deepseek-ai_deepseek_2024}
DeepSeek-AI, Xiao Bi, Deli Chen, Guanting Chen, Shanhuang Chen, Damai Dai, Chengqi Deng, Honghui Ding, Kai Dong, Qiushi Du, Zhe Fu, Huazuo Gao, Kaige Gao, Wenjun Gao, Ruiqi Ge, Kang Guan, Daya Guo, Jianzhong Guo, Guangbo Hao, Zhewen Hao, Ying He, Wenjie Hu, Panpan Huang, Erhang Li, Guowei Li, Jiashi Li, Yao Li, Y.~K. Li, Wenfeng Liang, Fangyun Lin, A.~X. Liu, Bo~Liu, Wen Liu, Xiaodong Liu, Xin Liu, Yiyuan Liu, Haoyu Lu, Shanghao Lu, Fuli Luo, Shirong Ma, Xiaotao Nie, Tian Pei, Yishi Piao, Junjie Qiu, Hui Qu, Tongzheng Ren, Zehui Ren, Chong Ruan, Zhangli Sha, Zhihong Shao, Junxiao Song, Xuecheng Su, Jingxiang Sun, Yaofeng Sun, Minghui Tang, Bingxuan Wang, Peiyi Wang, Shiyu Wang, Yaohui Wang, Yongji Wang, Tong Wu, Y.~Wu, Xin Xie, Zhenda Xie, Ziwei Xie, Yiliang Xiong, Hanwei Xu, R.~X. Xu, Yanhong Xu, Dejian Yang, Yuxiang You, Shuiping Yu, Xingkai Yu, B.~Zhang, Haowei Zhang, Lecong Zhang, Liyue Zhang, Mingchuan Zhang, Minghua Zhang, Wentao Zhang, Yichao Zhang, Chenggang Zhao, Yao Zhao, Shangyan Zhou, Shunfeng
  Zhou, Qihao Zhu, and Yuheng Zou.
\newblock {DeepSeek} {LLM}: {Scaling} {Open}-{Source} {Language} {Models} with {Longtermism}, January 2024.
\newblock URL \url{http://arxiv.org/abs/2401.02954}.
\newblock arXiv:2401.02954 [cs].

\bibitem[Dettmers et~al.(2022)Dettmers, Lewis, Belkada, and Zettlemoyer]{dettmers_gpt3int8_2022}
Tim Dettmers, Mike Lewis, Younes Belkada, and Luke Zettlemoyer.
\newblock {GPT3}.int8(): 8-bit {Matrix} {Multiplication} for {Transformers} at {Scale}.
\newblock \emph{Advances in Neural Information Processing Systems}, 35:\penalty0 30318--30332, December 2022.
\newblock URL \url{https://proceedings.neurips.cc/paper_files/paper/2022/hash/c3ba4962c05c49636d4c6206a97e9c8a-Abstract-Conference.html}.

\bibitem[Ding et~al.(2024)Ding, Cao, Xie, Fan, Wang, and Lu]{ding_lora-c_2024}
Chuntao Ding, Xu~Cao, Jianhang Xie, Linlin Fan, Shangguang Wang, and Zhichao Lu.
\newblock {LoRA}-{C}: {Parameter}-{Efficient} {Fine}-{Tuning} of {Robust} {CNN} for {IoT} {Devices}, November 2024.
\newblock URL \url{http://arxiv.org/abs/2410.16954}.
\newblock arXiv:2410.16954 [cs].

\bibitem[Eldan and Li(2023)]{eldan_tinystories_2023}
Ronen Eldan and Yuanzhi Li.
\newblock {TinyStories}: {How} {Small} {Can} {Language} {Models} {Be} and {Still} {Speak} {Coherent} {English}?, May 2023.
\newblock URL \url{http://arxiv.org/abs/2305.07759}.
\newblock arXiv:2305.07759 [cs].

\bibitem[Frantar et~al.(2023)Frantar, Ashkboos, Hoefler, and Alistarh]{frantar_gptq_2023}
Elias Frantar, Saleh Ashkboos, Torsten Hoefler, and Dan Alistarh.
\newblock {GPTQ}: {Accurate} {Post}-{Training} {Quantization} for {Generative} {Pre}-trained {Transformers}, March 2023.
\newblock URL \url{http://arxiv.org/abs/2210.17323}.
\newblock arXiv:2210.17323 [cs].

\bibitem[Guan et~al.(2022)Guan, Liu, and Huang]{guan_corpus_2022}
Jian Guan, Ziqi Liu, and Minlie Huang.
\newblock A {Corpus} for {Understanding} and {Generating} {Moral} {Stories}, April 2022.
\newblock URL \url{http://arxiv.org/abs/2204.09438}.
\newblock arXiv:2204.09438 [cs].

\bibitem[Hu et~al.(2021)Hu, Shen, Wallis, Allen-Zhu, Li, Wang, Wang, and Chen]{hu_lora_2021}
Edward~J. Hu, Yelong Shen, Phillip Wallis, Zeyuan Allen-Zhu, Yuanzhi Li, Shean Wang, Lu~Wang, and Weizhu Chen.
\newblock {LoRA}: {Low}-{Rank} {Adaptation} of {Large} {Language} {Models}, October 2021.
\newblock URL \url{http://arxiv.org/abs/2106.09685}.
\newblock arXiv:2106.09685 [cs].

\bibitem[Hu et~al.(2023)Hu, Yin, and Wan]{hu_exploring_2023}
Xinyu Hu, Xunjian Yin, and Xiaojun Wan.
\newblock Exploring {Context}-{Aware} {Evaluation} {Metrics} for {Machine} {Translation}.
\newblock In Houda Bouamor, Juan Pino, and Kalika Bali, editors, \emph{Findings of the {Association} for {Computational} {Linguistics}: {EMNLP} 2023}, pages 15291--15298, Singapore, December 2023. Association for Computational Linguistics.
\newblock \doi{10.18653/v1/2023.findings-emnlp.1021}.
\newblock URL \url{https://aclanthology.org/2023.findings-emnlp.1021/}.

\bibitem[Koehn(2005)]{koehn_europarl_2005}
Philipp Koehn.
\newblock Europarl: {A} {Parallel} {Corpus} for {Statistical} {Machine} {Translation}.
\newblock In \emph{The {Tenth} {Machine} {Translation} {Summit} {Proceedings} of {Conference}}, pages 79--86. International Association for Machine Translation, 2005.
\newblock URL \url{https://www.research.ed.ac.uk/en/publications/europarl-a-parallel-corpus-for-statistical-machine-translation}.

\bibitem[Kwon et~al.(2023)Kwon, Li, Zhuang, Sheng, Zheng, Yu, Gonzalez, Zhang, and Stoica]{kwon_efficient_2023}
Woosuk Kwon, Zhuohan Li, Siyuan Zhuang, Ying Sheng, Lianmin Zheng, Cody~Hao Yu, Joseph~E. Gonzalez, Hao Zhang, and Ion Stoica.
\newblock Efficient {Memory} {Management} for {Large} {Language} {Model} {Serving} with {PagedAttention}, September 2023.
\newblock URL \url{http://arxiv.org/abs/2309.06180}.
\newblock arXiv:2309.06180 [cs].

\bibitem[Lin et~al.(2024)Lin, Tang, Tang, Yang, Chen, Wang, Xiao, Dang, Gan, and Han]{lin_awq_2024}
Ji~Lin, Jiaming Tang, Haotian Tang, Shang Yang, Wei-Ming Chen, Wei-Chen Wang, Guangxuan Xiao, Xingyu Dang, Chuang Gan, and Song Han.
\newblock {AWQ}: {Activation}-aware {Weight} {Quantization} for {LLM} {Compression} and {Acceleration}, July 2024.
\newblock URL \url{http://arxiv.org/abs/2306.00978}.
\newblock arXiv:2306.00978 [cs].

\bibitem[Liu et~al.(2023)Liu, Iter, Xu, Wang, Xu, and Zhu]{liu_g-eval_2023}
Yang Liu, Dan Iter, Yichong Xu, Shuohang Wang, Ruochen Xu, and Chenguang Zhu.
\newblock G-{Eval}: {NLG} {Evaluation} using {GPT}-4 with {Better} {Human} {Alignment}, May 2023.
\newblock URL \url{http://arxiv.org/abs/2303.16634}.
\newblock arXiv:2303.16634 [cs].

\bibitem[Lommel et~al.(2014)Lommel, Uszkoreit, and Burchardt]{lommel_multidimensional_2014}
Arle Lommel, Hans Uszkoreit, and Aljoscha Burchardt.
\newblock Multidimensional {Quality} {Metrics} ({MQM}) : a {Framework} for {Declaring} and {Describing} {Translation} {Quality} {Metrics}.
\newblock \emph{Tradumàtica}, 1\penalty0 (12):\penalty0 0455--463, 2014.
\newblock ISSN 1578-7559.
\newblock \doi{10.5565/rev/tradumatica.77}.
\newblock URL \url{https://ddd.uab.cat/record/130144}.

\bibitem[Long et~al.(2024)Long, Wang, Xiao, Zhao, Ding, Chen, and Wang]{long_llms-driven_2024}
Lin Long, Rui Wang, Ruixuan Xiao, Junbo Zhao, Xiao Ding, Gang Chen, and Haobo Wang.
\newblock On {LLMs}-{Driven} {Synthetic} {Data} {Generation}, {Curation}, and {Evaluation}: {A} {Survey}, June 2024.
\newblock URL \url{http://arxiv.org/abs/2406.15126}.
\newblock arXiv:2406.15126 [cs].

\bibitem[Mao et~al.(2024)Mao, Ge, Fan, Xu, Mi, Hu, and Gao]{mao_survey_2024}
Yuren Mao, Yuhang Ge, Yijiang Fan, Wenyi Xu, Yu~Mi, Zhonghao Hu, and Yunjun Gao.
\newblock A survey on {LoRA} of large language models.
\newblock \emph{Frontiers of Computer Science}, 19\penalty0 (7):\penalty0 197605, December 2024.
\newblock ISSN 2095-2236.
\newblock \doi{10.1007/s11704-024-40663-9}.
\newblock URL \url{https://doi.org/10.1007/s11704-024-40663-9}.

\bibitem[Martins et~al.(2025)Martins, Alves, Fernandes, Guerreiro, Rei, Farajian, Klimaszewski, Alves, Pombal, Boizard, Faysse, Colombo, Yvon, Haddow, Souza, Birch, and Martins]{martins_eurollm-9b_2025}
Pedro~Henrique Martins, João Alves, Patrick Fernandes, Nuno~M. Guerreiro, Ricardo Rei, Amin Farajian, Mateusz Klimaszewski, Duarte~M. Alves, José Pombal, Nicolas Boizard, Manuel Faysse, Pierre Colombo, François Yvon, Barry Haddow, José G. C.~de Souza, Alexandra Birch, and André F.~T. Martins.
\newblock {EuroLLM}-{9B}: {Technical} {Report}, June 2025.
\newblock URL \url{http://arxiv.org/abs/2506.04079}.
\newblock arXiv:2506.04079 [cs].

\bibitem[Masala et~al.(2024)Masala, Ilie-Ablachim, Dima, Corlatescu, Zavelca, Olaru, Terian, Terian, Leordeanu, Velicu, Popescu, Dascalu, and Rebedea]{masala_vorbesti_2024}
Mihai Masala, Denis~C. Ilie-Ablachim, Alexandru Dima, Dragos Corlatescu, Miruna Zavelca, Ovio Olaru, Simina Terian, Andrei Terian, Marius Leordeanu, Horia Velicu, Marius Popescu, Mihai Dascalu, and Traian Rebedea.
\newblock "{Vorbeşti} {Româneşte}?" {A} {Recipe} to {Train} {Powerful} {Romanian} {LLMs} with {English} {Instructions}, October 2024.
\newblock URL \url{http://arxiv.org/abs/2406.18266}.
\newblock arXiv:2406.18266 [cs].

\bibitem[Nadas et~al.(2025{\natexlab{a}})Nadas, Diosan, Piscoran, and Tomescu]{nadas_tf1-en-3m_2025}
Mihai Nadas, Laura Diosan, Andrei Piscoran, and Andreea Tomescu.
\newblock {TF1}-{EN}-{3M}: {Three} {Million} {Synthetic} {Moral} {Fables} for {Training} {Small}, {Open} {Language} {Models}, April 2025{\natexlab{a}}.
\newblock URL \url{http://arxiv.org/abs/2504.20605}.
\newblock arXiv:2504.20605 [cs].

\bibitem[Nadas et~al.(2025{\natexlab{b}})Nadas, Diosan, and Tomescu]{nadas_synthetic_2025}
Mihai Nadas, Laura Diosan, and Andreea Tomescu.
\newblock Synthetic {Data} {Generation} {Using} {Large} {Language} {Models}: {Advances} in {Text} and {Code}, March 2025{\natexlab{b}}.
\newblock URL \url{http://arxiv.org/abs/2503.14023}.
\newblock arXiv:2503.14023 [cs].

\bibitem[Papineni et~al.(2002)Papineni, Roukos, Ward, and Zhu]{papineni_bleu_2002}
Kishore Papineni, Salim Roukos, Todd Ward, and Wei-Jing Zhu.
\newblock Bleu: a {Method} for {Automatic} {Evaluation} of {Machine} {Translation}.
\newblock In Pierre Isabelle, Eugene Charniak, and Dekang Lin, editors, \emph{Proceedings of the 40th {Annual} {Meeting} of the {Association} for {Computational} {Linguistics}}, pages 311--318, Philadelphia, Pennsylvania, USA, July 2002. Association for Computational Linguistics.
\newblock \doi{10.3115/1073083.1073135}.
\newblock URL \url{https://aclanthology.org/P02-1040/}.

\bibitem[Rei et~al.(2020)Rei, Stewart, Farinha, and Lavie]{rei_comet_2020}
Ricardo Rei, Craig Stewart, Ana~C. Farinha, and Alon Lavie.
\newblock {COMET}: {A} {Neural} {Framework} for {MT} {Evaluation}, October 2020.
\newblock URL \url{http://arxiv.org/abs/2009.09025}.
\newblock arXiv:2009.09025 [cs].

\bibitem[Sennrich et~al.(2016)Sennrich, Haddow, and Birch]{sennrich_improving_2016}
Rico Sennrich, Barry Haddow, and Alexandra Birch.
\newblock Improving {Neural} {Machine} {Translation} {Models} with {Monolingual} {Data}, June 2016.
\newblock URL \url{http://arxiv.org/abs/1511.06709}.
\newblock arXiv:1511.06709 [cs].

\bibitem[Shaw and Kurtz(2024)]{shaw_llm_2024}
Robert Shaw and Mark Kurtz.
\newblock {LLM} {Compressor} is here: {Faster} inference with {vLLM}, August 2024.
\newblock URL \url{https://developers.redhat.com/articles/2024/08/14/llm-compressor-here-faster-inference-vllm}.

\bibitem[Silva et~al.(2024)Silva, Srivastava, Moteu~Ngoli, Röder, Moussallem, and Ngonga~Ngomo]{silva_benchmarking_2024}
Ana Silva, Nikit Srivastava, Tatiana Moteu~Ngoli, Michael Röder, Diego Moussallem, and Axel-Cyrille Ngonga~Ngomo.
\newblock Benchmarking {Low}-{Resource} {Machine} {Translation} {Systems}.
\newblock In Atul~Kr. Ojha, Chao-hong Liu, Ekaterina Vylomova, Flammie Pirinen, Jade Abbott, Jonathan Washington, Nathaniel Oco, Valentin Malykh, Varvara Logacheva, and Xiaobing Zhao, editors, \emph{Proceedings of the {Seventh} {Workshop} on {Technologies} for {Machine} {Translation} of {Low}-{Resource} {Languages} ({LoResMT} 2024)}, pages 175--185, Bangkok, Thailand, August 2024. Association for Computational Linguistics.
\newblock \doi{10.18653/v1/2024.loresmt-1.18}.
\newblock URL \url{https://aclanthology.org/2024.loresmt-1.18/}.

\bibitem[Sulem et~al.(2018)Sulem, Abend, and Rappoport]{sulem_bleu_2018}
Elior Sulem, Omri Abend, and Ari Rappoport.
\newblock {BLEU} is {Not} {Suitable} for the {Evaluation} of {Text} {Simplification}, October 2018.
\newblock URL \url{http://arxiv.org/abs/1810.05995}.
\newblock arXiv:1810.05995 [cs].

\bibitem[Sun and Duh(2020)]{sun_clirmatrix_2020}
Shuo Sun and Kevin Duh.
\newblock {CLIRMatrix}: {A} massively large collection of bilingual and multilingual datasets for {Cross}-{Lingual} {Information} {Retrieval}.
\newblock In Bonnie Webber, Trevor Cohn, Yulan He, and Yang Liu, editors, \emph{Proceedings of the 2020 {Conference} on {Empirical} {Methods} in {Natural} {Language} {Processing} ({EMNLP})}, pages 4160--4170, Online, November 2020. Association for Computational Linguistics.
\newblock \doi{10.18653/v1/2020.emnlp-main.340}.
\newblock URL \url{https://aclanthology.org/2020.emnlp-main.340/}.

\bibitem[Team et~al.(2022)Team, Costa-jussà, Cross, Çelebi, Elbayad, Heafield, Heffernan, Kalbassi, Lam, Licht, Maillard, Sun, Wang, Wenzek, Youngblood, Akula, Barrault, Gonzalez, Hansanti, Hoffman, Jarrett, Sadagopan, Rowe, Spruit, Tran, Andrews, Ayan, Bhosale, Edunov, Fan, Gao, Goswami, Guzmán, Koehn, Mourachko, Ropers, Saleem, Schwenk, and Wang]{team_no_2022}
Nllb Team, Marta~R. Costa-jussà, James Cross, Onur Çelebi, Maha Elbayad, Kenneth Heafield, Kevin Heffernan, Elahe Kalbassi, Janice Lam, Daniel Licht, Jean Maillard, Anna Sun, Skyler Wang, Guillaume Wenzek, Al~Youngblood, Bapi Akula, Loic Barrault, Gabriel~Mejia Gonzalez, Prangthip Hansanti, John Hoffman, Semarley Jarrett, Kaushik~Ram Sadagopan, Dirk Rowe, Shannon Spruit, Chau Tran, Pierre Andrews, Necip~Fazil Ayan, Shruti Bhosale, Sergey Edunov, Angela Fan, Cynthia Gao, Vedanuj Goswami, Francisco Guzmán, Philipp Koehn, Alexandre Mourachko, Christophe Ropers, Safiyyah Saleem, Holger Schwenk, and Jeff Wang.
\newblock No {Language} {Left} {Behind}: {Scaling} {Human}-{Centered} {Machine} {Translation}, July 2022.
\newblock URL \url{https://arxiv.org/abs/2207.04672v3}.

\bibitem[Tiedemann(2012)]{tiedemann_parallel_2012}
Jorg Tiedemann.
\newblock Parallel {Data}, {Tools} and {Interfaces} in {OPUS}.
\newblock \emph{Lrec}, 2012:\penalty0 2214--2218, 2012.

\bibitem[Wang et~al.(2025)Wang, Hu, and Ali]{wang_maats_2025}
George Wang, Jiaqian Hu, and Safinah Ali.
\newblock {MAATS}: {A} {Multi}-{Agent} {Automated} {Translation} {System} {Based} on {MQM} {Evaluation}, August 2025.
\newblock URL \url{http://arxiv.org/abs/2505.14848}.
\newblock arXiv:2505.14848 [cs].

\bibitem[Wang et~al.(2024)Wang, Jain, Zhang, Ray, Kumar, and Athiwaratkun]{wang_reasoning_2024}
Junlin Wang, Siddhartha Jain, Dejiao Zhang, Baishakhi Ray, Varun Kumar, and Ben Athiwaratkun.
\newblock Reasoning in {Token} {Economies}: {Budget}-{Aware} {Evaluation} of {LLM} {Reasoning} {Strategies}, June 2024.
\newblock URL \url{http://arxiv.org/abs/2406.06461}.
\newblock arXiv:2406.06461 [cs].

\bibitem[Wang et~al.(2023)Wang, Kordi, Mishra, Liu, Smith, Khashabi, and Hajishirzi]{wang_self-instruct_2023_2}
Yizhong Wang, Yeganeh Kordi, Swaroop Mishra, Alisa Liu, Noah~A. Smith, Daniel Khashabi, and Hannaneh Hajishirzi.
\newblock Self-{Instruct}: {Aligning} {Language} {Models} with {Self}-{Generated} {Instructions}, May 2023.
\newblock URL \url{http://arxiv.org/abs/2212.10560}.
\newblock arXiv:2212.10560 [cs].

\bibitem[Weyssow et~al.(2025)Weyssow, Zhou, Kim, Lo, and Sahraoui]{weyssow_exploring_2025}
Martin Weyssow, Xin Zhou, Kisub Kim, David Lo, and Houari Sahraoui.
\newblock Exploring {Parameter}-{Efficient} {Fine}-{Tuning} {Techniques} for {Code} {Generation} with {Large} {Language} {Models}.
\newblock \emph{ACM Trans. Softw. Eng. Methodol.}, January 2025.
\newblock ISSN 1049-331X.
\newblock \doi{10.1145/3714461}.
\newblock URL \url{https://dl.acm.org/doi/10.1145/3714461}.
\newblock Just Accepted.

\bibitem[Xue et~al.(2021)Xue, Constant, Roberts, Kale, Al-Rfou, Siddhant, Barua, and Raffel]{xue_mt5_2021}
Linting Xue, Noah Constant, Adam Roberts, Mihir Kale, Rami Al-Rfou, Aditya Siddhant, Aditya Barua, and Colin Raffel.
\newblock {mT5}: {A} massively multilingual pre-trained text-to-text transformer, March 2021.
\newblock URL \url{http://arxiv.org/abs/2010.11934}.
\newblock arXiv:2010.11934 [cs] version: 3.

\bibitem[Yao et~al.(2024)Yao, Jiang, Bobinac, Yang, and Hu]{yao_benchmarking_2024}
Binwei Yao, Ming Jiang, Tara Bobinac, Diyi Yang, and Junjie Hu.
\newblock Benchmarking {Machine} {Translation} with {Cultural} {Awareness}, October 2024.
\newblock URL \url{http://arxiv.org/abs/2305.14328}.
\newblock arXiv:2305.14328 [cs].

\bibitem[Zhang et~al.(2025)Zhang, Zhao, Macken, and Eger]{zhang_litransproqa_2025}
Ran Zhang, Wei Zhao, Lieve Macken, and Steffen Eger.
\newblock {LiTransProQA}: an {LLM}-based {Literary} {Translation} evaluation metric with {Professional} {Question} {Answering}, May 2025.
\newblock URL \url{http://arxiv.org/abs/2505.05423}.
\newblock arXiv:2505.05423 [cs].

\bibitem[Zheng et~al.(2023)Zheng, Chiang, Sheng, Zhuang, Wu, Zhuang, Lin, Li, Li, Xing, Zhang, Gonzalez, and Stoica]{zheng_judging_2023}
Lianmin Zheng, Wei-Lin Chiang, Ying Sheng, Siyuan Zhuang, Zhanghao Wu, Yonghao Zhuang, Zi~Lin, Zhuohan Li, Dacheng Li, Eric~P. Xing, Hao Zhang, Joseph~E. Gonzalez, and Ion Stoica.
\newblock Judging {LLM}-as-a-{Judge} with {MT}-{Bench} and {Chatbot} {Arena}, December 2023.
\newblock URL \url{http://arxiv.org/abs/2306.05685}.
\newblock arXiv:2306.05685 [cs].

\bibitem[Wang et~al.(2023)Wang, Laubli, Sennrich, Simen, Knowles, and others]{wang_findings_2023}
Longyue Wang, Zhaopeng Tu, Siyou Liu, Shuming Shi, Philipp Koehn, Liting Zhou, and Andy Way.
\newblock Findings of the {WMT} 2023 {Shared} {Task} on {Discourse}-{Level} {Literary} {Translation}: {A} {Fresh} {Orb} in the {Cosmos} of {LLMs}.
\newblock In \emph{Proceedings of the Eighth Conference on Machine Translation}, pages 55--70, Singapore, December 2023. Association for Computational Linguistics.
\newblock URL \url{https://aclanthology.org/2023.wmt-1.4/}.

\bibitem[Karpinska and Iyyer(2023)]{karpinska_large_2023}
Marzena Karpinska and Mohit Iyyer.
\newblock Large {Language} {Models} {Effectively} {Leverage} {Document}-level {Context} for {Literary} {Translation}, but {Critical} {Errors} {Persist}.
\newblock In \emph{Proceedings of the Eighth Conference on Machine Translation}, pages 419--451, Singapore, December 2023. Association for Computational Linguistics.
\newblock URL \url{https://aclanthology.org/2023.wmt-1.41/}.

\bibitem[Zhang et~al.(2025)Zhang, Zhao, and Eger]{zhang_how_2025}
Ran Zhang, Wei Zhao, and Steffen Eger.
\newblock How good are {LLMs} for literary translation, really? {Literary} translation evaluation with humans and {LLMs}.
\newblock In \emph{Proceedings of the 2025 Conference of the Nations of the Americas Chapter of the Association for Computational Linguistics: Human Language Technologies (Volume 1: Long Papers)}, 2025. Association for Computational Linguistics.

\bibitem[Kocmi and Federmann(2023)]{kocmi_large_2023}
Tom Kocmi and Christian Federmann.
\newblock Large {Language} {Models} {Are} {State}-of-the-{Art} {Evaluators} of {Translation} {Quality}.
\newblock In \emph{Proceedings of the 24th Annual Conference of the European Association for Machine Translation}, pages 193--203, Tampere, Finland, June 2023. European Association for Machine Translation.
\newblock URL \url{https://aclanthology.org/2023.eamt-1.19/}.

\bibitem[Wataoka et~al.(2024)Wataoka, Ozaki, Takayama, and Yokota]{wataoka_selfpref_2024}
Koki Wataoka, Tsubasa Ozaki, Daisuke Takayama, and Rio Yokota.
\newblock Self-{Preference} {Bias} in {LLM}-as-a-{Judge}, October 2024.
\newblock URL \url{http://arxiv.org/abs/2410.21819}.
\newblock arXiv:2410.21819 [cs].

\end{thebibliography}

\appendix
\section*{Appendix A. Cost Calculation and Hardware Configurations}
\label{sec:appendix:costs}

\subsection*{A.1 Cost Estimation Methodology}

Table~\ref{tab:cost-details} details the assumptions used to compute cost estimates in Section~\ref{sec:costs}. 
We model cost as a function of input tokens ($T_{in}$), output tokens ($T_{out}$), and API pricing per million tokens.
For proprietary reasoning models (e.g., GPT-o3), we assume hidden reasoning tokens are billed as output, with 
a ``medium reasoning'' setting (reasoning tokens $\approx$ visible output). 

\begin{equation}
\text{Total Cost} = \bigg( \frac{T_{in}}{10^6} \cdot P_{in} \bigg) + \bigg( \frac{T_{out} + T_{reason}}{10^6} \cdot P_{out} \bigg),
\end{equation}
where $P_{in}$ and $P_{out}$ denote input/output pricing (\$/million tokens) and $T_{reason}$ are hidden tokens.

\begin{table}[!htbp]
\centering
\footnotesize
\begin{tabular}{lrrr}
\toprule
Model & $P_{in}$ (\$/M) & $P_{out}$ (\$/M) & Notes \\
\midrule
GPT-4.1 & 2.00 & 8.00 & Standard API pricing (Aug 2025) \\
GPT-4.1-mini & 0.40 & 1.60 & Lower capacity, same billing rules \\
GPT-o3 & 2.00 & 8.00 & Reasoning tokens billed as output \\
GPT-o3-mini & 1.10 & 4.40 & Same reasoning token policy \\
DeepL API Pro & -- & -- & Flat monthly + per-character rate, converted to tokens \\
TF2 models (ours) & -- & -- & Rented GPUs ($\approx 340$\$ for 3M fables) \\
\bottomrule
\end{tabular}
\caption{Per-million-token pricing assumptions used in cost calculations.}
\label{tab:cost-details}
\end{table}

\subsection*{A.2 Hardware Configurations for TF2 Models}

Table~\ref{tab:hardware} lists the compute environments used for training and inference. 
We relied exclusively on commodity GPUs (cloud or local) with support for FP16/bfloat16 mixed precision 
and 8-bit quantization (W8A8). All inference experiments were executed via the \texttt{OpenRouter} API, 
which provided unified access to both open-weight and proprietary models.

\begin{table}[!htbp]
\centering
\footnotesize
\begin{tabular}{l l l l}
\toprule
Stage & Hardware & Runtime & Notes \\
\midrule
Fine-tuning TF2-1B & 1 $\times$ l40s & 1h & LoRA adapters, FP16 \\
Fine-tuning TF2-4B & 1 $\times$ l40s & 2-3h & Gradient accumulation, FP16 \\
Fine-tuning TF2-12B & 1 $\times$ h100 & 2h & FP16, early stopping \\
Inference (3M fables) & 8 $\times$ h100 & $\sim$31h & sfcompute clusters, vLLM endpoints \\
Quantization & CPU + GPU mix & $<$1h per model & W8A8 compression with \texttt{llmcompressor} \\
\bottomrule
\end{tabular}
\caption{Hardware setups for training and inference. Runtime values are approximate wall-clock time.}
\label{tab:hardware}
\end{table}

\subsection*{A.3 Energy and Cost of Local Compute}

To approximate the cost of our TF2 pipeline, we base calculations on the provider we used (\texttt{sfcompute}):
\begin{itemize}
    \item \textbf{GPU rental:} \$1.35 per GPU-hour; our 8-GPU node is billed at \$10.80 per \emph{cluster}-hour.
    \item \textbf{Electricity:} Not applicable (cloud rental includes energy). For on-prem scenarios only, a rough estimate is 300--450\,W/GPU at \$0.14/kWh (\(\approx\$0.04\)--\(\$0.06\) per GPU-hour).
\end{itemize}

The resulting rental cost is
\[
C_{\text{rental}}~[\$] \;=\; 10.8 \times H_{\text{cluster}},
\]
where \(H_{\text{cluster}}\) is the logged wall-clock cluster time (hours). For reference, 24, 32, and 40 cluster-hours correspond to \$259.2, \$345.6, and \$432, respectively. Storage and network costs were negligible in our runs.

\subsection*{A.4 Software Environment}

Experiments were conducted in the following software stack:
\begin{itemize}
    \item \textbf{OS:} Manjaro + Linux 6.12
    \item \textbf{Frameworks:} PyTorch 2.7.1, Hugging Face Transformers 4.54.0, PEFT 0.11
    \item \textbf{Inference:} vLLM 0.7.3, FlashAttention
    \item \textbf{Quantization:} \texttt{llmcompressor} (Aug 2024 release)
    \item \textbf{Deployment artifacts:} GGUF and Safetensors formats
\end{itemize}

This environment ensures reproducibility and efficient deployment across both local and cloud hardware.

\appendix
\section*{Appendix B. Qualitative Translation Examples (Five Cases)}
\label{ap:examples}

We present five representative problem cases. For each, we show excerpts from the \emph{Original (EN)}, the \emph{Gemma-3-12B (untuned)} output, and the \emph{TF2-12B (fine-tuned)} output. We highlight the key segments where fine-tuning clearly improves lexical choice, species fidelity, and idiomatic Romanian.

\begin{table}[!htbp]
\centering
\footnotesize
\begin{tabular}{p{0.30\linewidth} p{0.33\linewidth} p{0.33\linewidth}}
\toprule
\textbf{Original (EN)} & \textbf{Gemma-3-12B (untuned)} & \textbf{TF2-12B (fine-tuned)} \\
\midrule

\textit{“A \textbf{Greedy Skunk} loved to explore the temple, sniffing out shiny trinkets… As it gazed into the mirror, it saw a \textbf{kind-hearted Skunk} staring back. ‘I’m special because I’m the \textbf{best treasure hunter}!’”} & 
\textit{„Un \textbf{Fumeg de lăcomie} adora să exploreze templul… un \textbf{Fumeg frumos și cu inimă bună} îl privea… ‘Sunt special pentru că sunt cel mai bun \textbf{căutător de comori}!’”} &
\textit{„\textbf{Sconcsul Lacom} iubea să exploreze templul… un \textbf{Sconcs frumos și bun la suflet} îl privea… ‘Sunt special pentru că sunt cel mai bun \textbf{vânător de comori}!’”} \\
\midrule

\textit{“In a sleepy village… a kind and gentle \textbf{cheetah} lived. When a hungry lion appeared, she shared food and comforted him. Together with her friends, they decided to \textbf{plant a strong tree} to symbolize their friendship… helping others earns \textbf{lasting loyalty}.”} &
\textit{„…trăia o \textbf{ceată blândă și bună}. Când un leu flămând a apărut, ceata i-a împărțit mâncare… au decis să planteze un copac puternic… ajutorul oferit altora aduce o \textbf{loialitate de durată}.”} &
\textit{„…trăia o \textbf{ghepardă blândă și bună}. Când un leu flămând a apărut, gheparda și-a împărțit hrana… Au hotărât să planteze un copac puternic, \textbf{simbol al noii prietenii}… a-i ajuta pe ceilalți aduce \textbf{loialitate de durată}.”} \\
\midrule

\textit{“\textbf{The Hippopotamus’ Hidden Treasure}… She met a wise tortoise by the river, who told her, ‘\textbf{Sharing is caring}.’ With her friends’ help, she found the imposter and learned that wise counsel can \textbf{light the way}.”} &
\textit{„\textbf{Iepurașul și Comoara Ascunsă}… a întâlnit pe \textbf{un țestoasă} bătrân… ‘\textbf{Ampărtășirea este dragoste}.’… a învățat că sfatul înțelept poate \textbf{lumina drumul}.”} &
\textit{„\textbf{Comoara ascunsă a hipopotamului}… a întâlnit o \textbf{broască țestoasă} bătrână… ‘\textbf{A împărți înseamnă a-ți păsa}.’… a învățat că sfatul înțelept poate \textbf{lumina calea}.”} \\
\midrule

\textit{“In a sun-kissed flower field, a clever \textbf{skunk} named Stinky fell in love with a \textbf{butterfly}, Pretty. But her best friend, a grumpy old \textbf{owl}, disliked him. When they saw their \textbf{reflections} in the stream, they realized, ‘We’ve been judging each other wrong.’”} &
\textit{„Într-o luncă plină de flori, un \textbf{pui de prepeliță} șiret… s-a îndrăgostit de o \textbf{libelulă}… prietena ei, un \textbf{bufnițel} morocănos… au văzut reflexiile în \textbf{apa limpezi}… ‘Ne-am judecat greșit,’ spuse Miron.”} &
\textit{„Într-un câmp de flori scăldat de soare, un \textbf{sconcs} isteț… s-a îndrăgostit de un \textbf{fluture}… prietena ei, o \textbf{bufniță} bătrână și morocănoasă… și-au văzut \textbf{propriile reflexii}… ‘Ne-am judecat greșit,’ spuse Stinky.”} \\
\midrule

\textit{“In a forgotten shipyard, a greedy \textbf{puma} lived. As a storm flooded the docks, a wise old \textbf{owl} offered help. Together they escaped and he learned: ‘That’s the power of \textbf{teamwork}.’”} &
\textit{„Într-\textbf{o șantier naval} uitată, trăia un \textbf{pumnă malign}… o \textbf{bufnița zâmbi}: ‘Aceasta este puterea \textbf{muncii în echipă}.’”} &
\textit{„Într-un șantier naval uitat, trăia un \textbf{puma} lacom… \textbf{Bufnița a zâmbit}: ‘Aceasta este puterea \textbf{muncii în echipă}.’… și împreună \textbf{s-au tras la adăpost}.”} \\
\bottomrule
\end{tabular}
\caption{Excerpts from five difficult fables. TF2-12B consistently preserves correct species and natural idioms, while Gemma-3-12B often introduces species corruption, typos, or ungrammatical forms.}
\end{table}

\paragraph{Observations.}  
These cases highlight how Gemma-3-12B often mistranslates the main species (skunk→Fumeg, cheetah→ceată, hippo→iepuraș, skunk→prepeliță/libelulă, puma→pumnă). TF2-12B restores fidelity, grammaticality, and idiomatic Romanian, producing fluent, age-appropriate texts.

\end{document}